\documentclass{article}

\usepackage{arxiv}

\usepackage[utf8]{inputenc} 
\usepackage[T1]{fontenc}    
\usepackage{hyperref}       
\usepackage{url}            
\usepackage{booktabs}       
\usepackage{amsfonts}       
\usepackage{nicefrac}       
\usepackage{microtype}      
\usepackage{lipsum}		
\usepackage{graphicx}
\usepackage{doi}

\usepackage{graphicx}
\usepackage{array}
\usepackage{algorithm}
\usepackage{epstopdf}
\usepackage{algpseudocode}
\usepackage{multirow}
\usepackage{lscape}
\usepackage{hyperref}
\usepackage{url}
\usepackage{longtable}
\usepackage{color, colortbl}
\usepackage{amsmath}
\usepackage{subcaption}
\usepackage{multirow}
\usepackage{siunitx}%
\usepackage{longtable}

\usepackage{tabularx}
\usepackage{mwe}
\usepackage{graphbox} 

\title{HCR-Net: A deep learning based script independent handwritten character recognition network}

\author{
        Vinod Kumar Chauhan \\
	Department of Engineering Science,\\
	University of Oxford UK\\
	\texttt{vinod.kumar@eng.ox.ac.uk} \\
	\AND
	Sukhdeep Singh \\
	D.M. College, Panjab University, Chandigarh,\\
        Moga, Punjab, India \\
        \texttt{sukha13@ymail.com}\\
        \AND
        Anuj Sharma\thanks{Correspondence to: anujs@pu.ac.in}\\
        Department of Computer Science and Applications,\\
        Panjab University, Chandigarh, India\\
        \texttt{anujs@pu.ac.in} \\
        Homepage: \url{https://anuj-sharma.in}
}

\begin{document}
\maketitle

\begin{abstract}
    Handwritten character recognition (HCR) remains a challenging pattern recognition problem despite decades of research, and lacks research on script independent recognition techniques. {\color{black}This is mainly because of similar character structures, different handwriting styles, diverse scripts, handcrafted feature extraction techniques, unavailability of data and code, and the development of script-specific deep learning techniques.
    To address these limitations, we have proposed a script independent deep learning network for HCR research, called HCR-Net, that sets a new research direction for the field. HCR-Net is based on a novel transfer learning approach for HCR, which \textit{partly utilizes} feature extraction layers of a pre-trained network.} Due to transfer learning and image augmentation, HCR-Net provides faster and computationally efficient training, better performance and generalizations, and can work with small datasets.   
    HCR-Net is extensively evaluated on 40 publicly available datasets of Bangla, Punjabi, Hindi, English, Swedish, Urdu, Farsi, Tibetan, Kannada, Malayalam, Telugu, Marathi, Nepali and Arabic languages, and established 26 new benchmark results while performed close to the best results in the rest cases. HCR-Net showed performance improvements up to 11\% against the existing results and achieved a fast convergence rate showing up to 99\% of final performance in the very first epoch. HCR-Net significantly outperformed the state-of-the-art transfer learning techniques and also reduced the number of trainable parameters by 34\% as compared with the corresponding pre-trained network. To facilitate reproducibility and further advancements of HCR research, the complete code is publicly released at \url{https://github.com/jmdvinodjmd/HCR-Net}.
\end{abstract}


\section{Introduction}
\label{sec_intro}
{\color{black}Handwritten character recognition (HCR) is a widely studied and an important pattern recognition problem, e.g., \cite{Granlund1972,al2023threshold,hamida2023cursive,Lin2023}. It has a variety of useful applications, including, developing applications to help visually impaired people, classroom teaching, digitizing ancient documents, converting handwritten notes on tablets to text \cite{Singh20171,singh2021new}, writer age estimation \cite{huang2023writer}, gender identification \cite{dargan2023handwriting}, disease detection \cite{zhao2023significantly}, reading doctor prescriptions and in the automated processing of forms in railways, post-offices and government offices \cite{Lam1988,basu2010novel}.}

Handwriting recognition, depending upon the type of data as online and offline, can be broadly classified into two categories \cite{Singh20171,Singh2019OHG} and so need different techniques for recognition. Online data consists of a series of points $(x,y)$ where $x$ and $y$ are coordinates, recorded between pen down and pen up on a digital surface. On the other hand, offline data consists of images of characters. Online data can be converted to offline and then offline techniques can be applied to online data which can sometimes give better results, e.g., \cite{Singh2021}. So, this paper focuses only on offline HCR but presents examples of online handwriting datasets IAPR-11 and UJIPenChars, which are first converted to images and then the proposed recognition technique is applied.

\begin{figure}[htb!]
	\centering
	\includegraphics[width=0.9\linewidth]{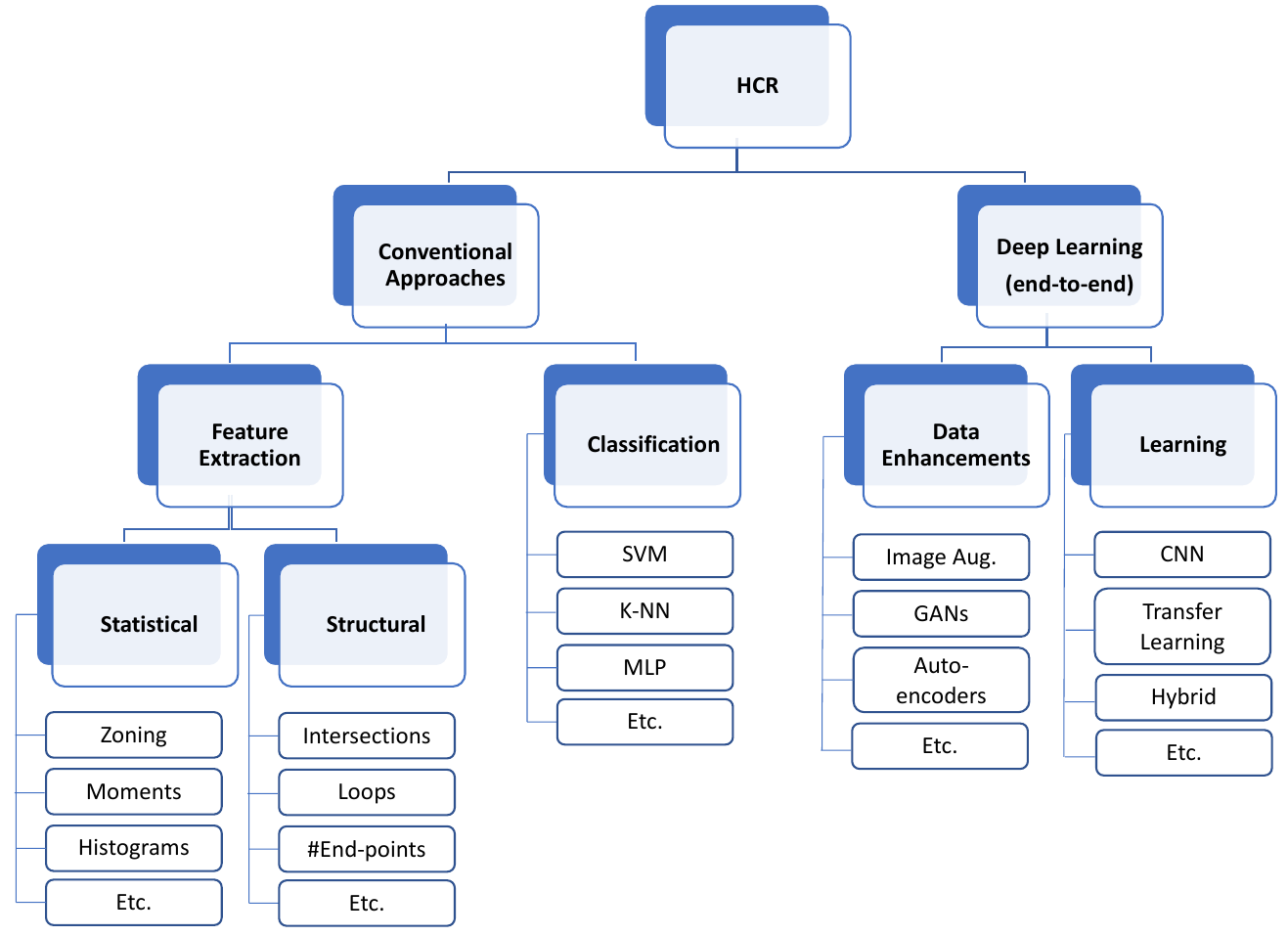}
	\caption{Handwritten character recognition research}
	\label{fig_HCR}
\end{figure}

Based on research approaches used for HCR, the research can be broadly classified into two categories: conventional approaches and deep learning based approaches, as shown in Fig.~\ref{fig_HCR} and discussed in detail in Section~\ref{sec_literature}. The conventional approaches focus on the development of feature extraction techniques and classifiers for recognition. Feature extraction involves finding distinct elements of characters which help to distinguish them from others. Generally, feature extraction develops handcrafted features based on the morphological or structural appearance of characters, which are domain/script specific and are not always available. Feature extraction can use statistical features, like zoning, histograms, and moments etc., or structural features, like the number of loops, intersections, and endpoints etc. \cite{Das2012,Das2012b,Roy2017,singh2021new}. Support vector machine (SVM), hidden Markov model (HMM), decision trees, k-nearest neighbour (k-NN) and multi-layer perceptrons (MLP) are widely used for recognition under the conventional approaches \cite{Das2014,Singh2016b,Hijam2021}. Unlike conventional approaches, deep learning provides end-to-end learning, i.e., performs both automated feature extraction and recognition. The recent success of deep learning, especially convolutional neural networks (CNN), also referred to as ConvNets, has shown impressive results in HCR and has become the first choice of the researchers, e.g., \cite{Singh20171,mukhoti2020handwritten,Guha2020}.

HCR has been studied for a few decades \cite{Granlund1972} but it is still an unsolved challenging learning problem in pattern recognition \cite{porwal2022advances,inunganbi2023systematic}. This is mainly because of the following reasons: (i) every person has their own writing style and no two persons' handwriting match, and that's why it is considered as a unique biometric identifier of a person, (ii) most of the scripts have characters which are similar in their structure, e.g., `ta' and `na' in Devanagari look similar, as shown in Sub-figs.~\ref{subfig_miss_classification_4} and \ref{subfig_miss_classification_5}, (iii) noisy data due to errors in the recording process, such as unwanted marks in the background of characters or errors in cropping images, e.g., in Sub-fig.~\ref{subfig_miss_classification_3} `digit\_9' is not cropped properly, (iv) bad handwriting, e.g., in Sub-fig.~\ref{subfig_miss_classification_2} `digit\_7' appears as `digit\_0', (v) recursive nature of scripts, e.g., Bangla, and (vi) unavailability of public datasets and code-repositories to reproduce and extend the existing results.

For some scripts, like Chinese, Japanese and Latin, there has been extensive research \cite{Lecun1998,Melnyk2020,Li2018,GAN2023109317,majid2022character}. However, for some other scripts, such as Gurmukhi, the research is still in its infancy. Since the research on individual scripts is considered an open problem and is ongoing \cite{Singh2021}, so there is little research on multi-script models. This is due to the following reasons: (i) the conventional research was focused on handcrafted features which are domain/script specific and are not always available, (ii) a large variety of scripts and their diversity, (iii) unavailability of datasets and code-repositories for extending and for reproducibility of results etc., and (iv) the existing deep learning techniques are developed for specific scripts only.

{\color{black}From the above discussion, it is clear that HCR remains a challenging pattern recognition problem and lacks the development of script independent recognition techniques. Deep learning offers end-to-end learning solutions for HCR, however, the existing deep learning techniques are customized to specific scripts/datasets, are computationally expensive and also lack reproducible code. Thus, the objective of the study is to develop a script independent, computationally efficient, faster, robust, publicly available and reproducible script independent deep learning technique for HCR.}

{\color{black}Building on the recent success and capability of end-to-end learning of deep learning, and the availability of publicly available datasets, this paper proposes the first script independent deep convolutional network for HCR, called HCR-Net. The proposed network is a script independent technique as it is not dependent on script-specific handcrafted features which might not be available for all the scripts. HCR-Net is based on a novel transfer learning approach for HCR, which \textit{partly utilizes} feature extraction layers of a pre-trained network, unlike the existing techniques \cite{Guha2020} that use the entire feature extraction layers. The proposed transfer learning approach is based on our hypothesis that the HCR task is simpler than those for which pre-trained models, like VGG16 \cite{VGG16} are developed, so the HCR does not need complex models.} HCR-Net has been extensively evaluated on 40 publicly available datasets and has established several new benchmarks, as discussed in Section~\ref{sec_experiments}.

The key contributions of the paper are summarized below.
\begin{itemize}
    \item[(a)] {\color{black}This paper proposes the first script independent novel deep convolutional network for end-to-end HCR, called HCR-Net, and sets a new research direction for the HCR field.}
    \item[(b)] HCR-Net develops a novel transfer learning approach for HCR research by partly utilizing feature extraction layers of a pre-trained VGG16 to initialize some of its lower layers, unlike the existing research which utilizes all feature extraction layers of the pre-trained models. Transfer learning along with image augmentation helps HCR-Net in faster, computationally efficient and robust learning and learning even on trivial datasets as compared with CNN models developed from scratch.
    \item[(c)] HCR-Net is extensively evaluated on 40 publicly available datasets of Bangla, Punjabi, Hindi, English, Swedish, Urdu, Farsi, Tibetan, Kannada, Malayalam, Telugu, Marathi, Nepali and Arabic languages, and established 26 new benchmark results while performing very close to the best results in the rest cases. HCR-Net showed performance improvements up to 11\% against the existing results. HCR-Net achieved a fast convergence rate and showed up to 99\% of final performance in the very first epoch. HCR-Net also significantly (p-value=0.00099 using Student's t-test) outperformed the existing transfer learning techniques and showed a 34\% reduction in the number of trainable parameters as compared with the corresponding pre-trained network.
    \item[(d)] For reproducibility and advancement of the HCR research, the complete code is released at: \url{https://github.com/jmdvinodjmd/HCR-Net}.
\end{itemize}

\paragraph{Organization of rest of the paper:} Section~\ref{sec_literature} presents a literature review and discusses conventional approaches and recent deep learning based approaches for HCR. Section~\ref{sec_HCRNet} discusses HCR-Net and Section~\ref{sec_experiments} presents experimental results on different scripts. Finally, Section~\ref{sec_conclusion} concludes the paper.

\section{Literature review}
\label{sec_literature}
In this section, literature on HCR is briefly discussed which can be broadly classified into two categories, conventional approaches and deep learning based approaches, as depicted in Fig.~\ref{fig_HCR} and discussed in the following subsections.

\subsection{Conventional approaches}
\label{subsec_literature_conventional}
HCR field has been studied extensively for more than five decades \cite{Granlund1972,Duerr1980,Lam1988,Lecun1998,Bhattacharya2005,Das2012,Acharya2015,Singh2016b,Sarkhel2017,Lincy2021,singh2023indic,muthureka2023improved,prijatelj2023novelty}. Earlier, the focus of research was mainly on developing feature extraction techniques and applying different classification techniques for recognition. Feature extraction is the process of finding key features which can distinguish different classes correctly and is a critical factor for the performance of machine learning models. Feature extraction can be further broadly classified into statistical and structural feature extraction techniques. Statistical feature extraction considers features based on pixel distribution in an image, e.g., histograms, zoning and moments. But structural feature extraction techniques consider features based on the structure of characters, such as loops, intersections, and number of endpoints. On the other hand, classification techniques are machine learning tools which learn to classify/recognize a script from a given feature/dataset. For example, SVM (for more details refer to \cite{Chauhan2019}), k-NN and MLP are the most widely used classifiers in the conventional approaches \cite{Das2012,Das2012b,Das2014,Hijam2021,Roy2017,Gupta2019}. A few representative conventional approaches are discussed below.

\cite{Granlund1972} proposed a Fourier transformation-based feature extraction along with a non-optimized decision method for the recognition of handwritten characters.
\cite{Lam1988} developed a system for the recognition of unconstrained handwritten digits using feature extraction based on geometric primitives containing topological information such as convex polygons and line segments, with a relaxation matching classifier.
\cite{Pal2000} proposed a novel feature extraction method based on the concept of water overflow from a reservoir as well as statistical and topological features along with a tree-based classifier for unconstrained offline handwritten Bangla numerals.
\cite{Bhattacharya2009} used wavelet-based multi-resolution features with multi-layer perceptron classifiers for digit recognition.
\cite{Das2012} proposed genetic algorithm (GA), simulated annealing and hill climbing techniques to sample regions to select local features. They used an SVM classifier for handwritten digit recognition.
\cite{Das2012b} proposed principal component analysis (PCA), modular PCA and quad-tree-based hierarchically derived longest-run features with SVM for recognition of numerals of Devanagari, Telugu, Bangla, Latin and Arabic scripts.
\cite{Das2014} presented a benchmark offline dataset of isolated handwritten Bangla compound characters, called CMATERdb~3.1.3.3. The recognition was performed using quad-tree-based features with SVM.
\cite{Ghosh2020} studied a multi-script numeral recognition for Bangla, Arabic, Telugu, Nepali, Assamese, Gurmukhi, Latin and Devanagari scripts. They used a histogram of oriented pixel positions and point-light source-based shadow feature extractors with k-NN, random forest, MLP, simple logistic and sequential minimal optimization as classifiers.

\subsection{Deep learning approaches}
\label{subsec_literature_DL}
The recent success of deep learning models, especially CNN, has revolutionized the artificial intelligence world and has found applications in different fields like, image processing, computer vision, healthcare and natural language processing \cite{Krizhevsky2012,Szegedy2015going,Sufian2020,chauhan2022coper,chauhan2023adversarial,chauhan2022continuous,chauhan2023brief,singh2023language,chauhan2023dynamic}. The success of deep learning models can be attributed, mainly to the advancements in the hardware technology, new optimization algorithms and availability of large number of data sources. CNN has shifted the paradigm from handcrafted features to automated features learned directly from the input images. CNN also outperforms all other machine learning techniques for HCR and has become the choice of researchers \cite{Deore2020,Guha2020,Manjusha2019,Roy2017,Rao2018,Ali2020pioneer}. However, the main limitations of CNN are that they need large amounts of data, great computing resources and large training time if trained from scratch. These limitations are overcome with the use of image augmentation and transfer learning techniques.
The CNN are the state-of-art for HCR research and a few important studies are discussed below.

\cite{Kim2015} proposed a CNN-based architecture for Hangul HCR and reported results of 95.96\% and 92.92\% on SERI95a and PE92 datasets, respectively.
\cite{Roy2017} employed a layer-wise training of CNN-based architecture for isolated Bangla compound character recognition. The proposed model was reported to outperform conventional shallow models, like SVM, as well as regular CNN.
\cite{Li2018} proposed a cascaded CNN with weighted average pooling for reducing the number of parameters for Chinese HCR. They reported 97.1\% results on the ICDAR-2013 dataset.
\cite{Manjusha2018} also proposed a CNN-based architecture utilizing scattering transform-based wavelet filters in the first convolutional layer for Malayalam HCR.
\cite{Rao2018} designed a lighter multi-channel residual CNN network (similar to GoogLeNet \cite{Szegedy2015going}) for handwritten digit recognition and reported results on mnist and SVHN datasets.
\cite{Kavitha2019} proposed a CNN-based architecture for offline Tamil HCR on HP Labs India dataset and achieved an accuracy of 97.7\%.
\cite{Keserwani2019} developed a CNN-based architecture for low-memory GPU for offline Bangla HCR. They used spatial pyramid pooling and fusion of features from different CNN layers.
\cite{Guha2020} proposed DevNet, a CNN-based architecture with five convolutional layers followed by max pooling, one fully connected layer and one fully connected layer as output, for Devanagari HCR.
\cite{Melnyk2020} presented a high-performance CNN-based architecture using global weighted output average pooling to calculate class activation maps for offline Chinese HCR.
\cite{Hijam2021} introduced the Meitei Mayek (Manipuri script) handwritten character dataset. They reported results using handcrafted features such as HOG and discrete wavelet transform (DWT), and image pixel intensities with random forest, k-NN, SVM and also using CNN-based architecture. CNN model provided benchmark results of 95.56\%.
\cite{Lincy2021} used a CNN-based architecture which uses a self-adaptive lion algorithm for fine-tuning fully connected layers and weights for Tamil HCR.
\cite{Inunganbi2021} proposed a three-channel CNN architecture using gradient direction, gradient magnitude and greyscale images for Meitei Mayek HCR.

Transfer learning is very successful in working with small datasets, including HCR \cite{Pramanik2018,Guha2020,Deore2020,Singh2021}. For example, \cite{Deore2020} used fine-tuned VGG16 in two stages for recognition of Devanagari and Bangla scripts. \cite{Pramanik2018} also used fine-tuning of pre-trained AlexNet and VGG16 on some Indic scripts.
Image augmentation, generative adversarial networks (GANs) and auto-encoders also help to work with limited datasets \cite{Deore2020,Chowdhury2019,Sufian2020,Kaur2020handwritten,Ali2020pioneer,Kong2019}.
Image augmentation artificially expands datasets by using operations, such as translation, flip, rotation, shear and zoom. on the input images. This helps in developing a robust classifier with limited datasets because the model is trained on the modified variants of the training images, e.g., \cite{Deore2020,Chowdhury2019,Sufian2020}. 
GANs are deep neural networks which are used to generate new, synthetic data, similar to real data. For example, \cite{Kaur2020handwritten} used GANs for Devanagari handwritten character generation. Auto-encoders are also deep neural networks which are used to learn compact representations of the data, like PCA and also for generating synthetic data, e.g., \cite{Ali2020pioneer} used deep encoder and CNN for recognition of handwritten Urdu characters.

In addition, a hybrid of conventional and deep learning approaches is also developed for HCR. For example, \cite{Maitra2015} used LeNet-5 for feature extraction and SVM as a classifier for recognition of Bangla, Devanagari, Latin, Oriya and Telugu. \cite{Manjusha2019} used scattering CNN with SVM for Malayalam HCR. \cite{Sarkhel2017} proposed a multi-column multi-scale CNN architecture based on a multi-scale deep quad tree-based feature extraction and used SVM as a classifier. They reported their results with Bangla, Tamil, Telugu, Hindi and Urdu scripts.

\par {\color{black}Thus, from the brief literature review, we find that conventional approaches are not suitable for script independent HCR due to the use of handcrafted features or manually designed features based on morphological or structural appearance, which might not be available all the time. On the other hand, recent developments in deep learning approaches due to their end-to-end learning approach are suitable but are studied a little for multi-script HCR. Moreover, the existing deep learning techniques are computationally expensive, lack reproducible code and are developed specifically for some scripts and may not work on other scripts.}

\section{HCR-Net}
\label{sec_HCRNet}
{\color{black}In this section, we discuss the architecture of HCR-Net, the contribution of transfer learning and image augmentation to HCR-Net, and the discussion and analysis of the two phase training of the proposed network.}

\subsection{Proposed architecture}
\label{subsec_proposed_model}
{\color{black}HCR-Net is a CNN-based end-to-end architecture for offline HCR whose lower and middle layers act as feature extractors and upper layers act as classifiers. HCR-Net partly utilizes the feature extraction layers of a pre-trained VGG16 network (as shown in Fig.~\ref{fig_HCRNet_architecture}) for initializing some of its lower layers, and trains in two phases. It is based on the hypothesis that the HCR is a relatively simple task as compared to ImageNet tasks on which most of the pre-trained deep learning networks are developed, e.g., VGG16 was originally trained on ImageNet with 14 million images and 1000 classes \cite{VGG16}. So, the use of only some of the lower layers of pre-trained models could be sufficient and could give better results for the HCR, and this is supported by our empirical results. This is also the reason for using VGG16 in HCR-Net and not using complex and powerful architectures, such as ResNet, DenseNet and Inception that have a large number of layers but are not useful for HCR.

\begin{figure}[htb!]
	\centering
	\includegraphics[width=\linewidth]{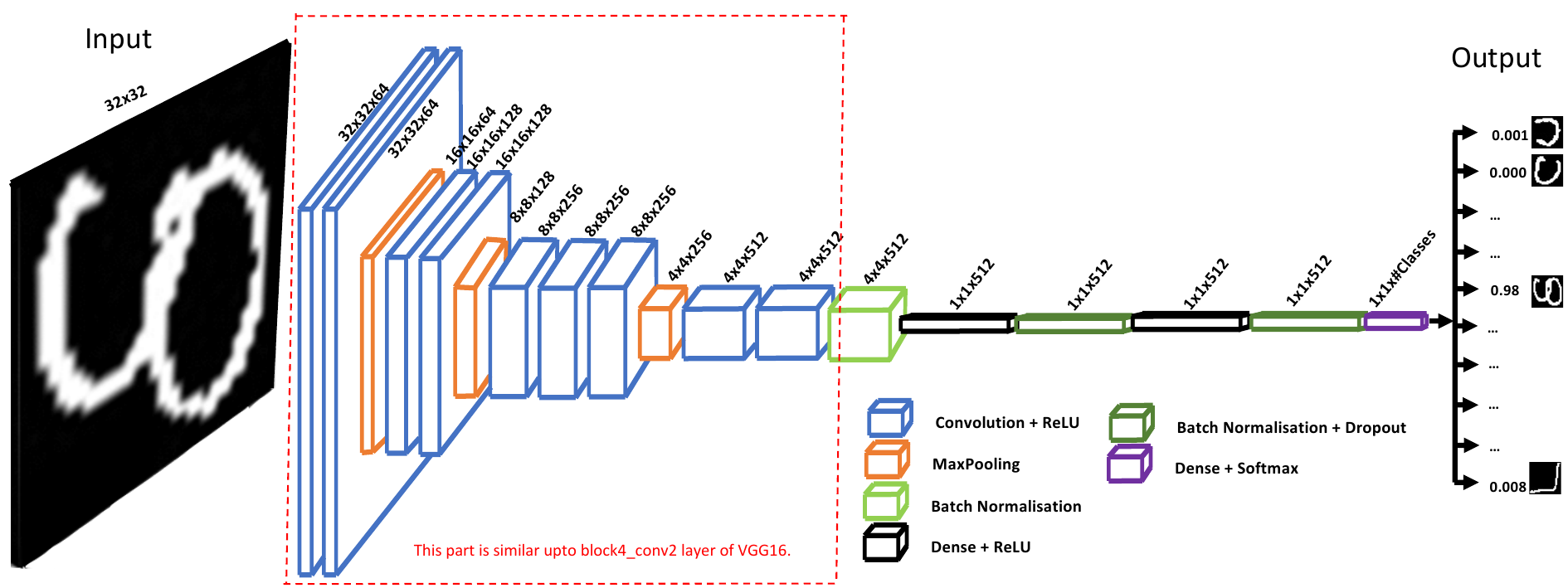}
	\caption{Proposed architecture of HCR-Net}
	\label{fig_HCRNet_architecture}
\end{figure}

Fig.~\ref{fig_HCRNet_architecture} presents the architecture of HCR-Net. It takes an input as a greyscale image of $32\times32$ pixels and produces output as class probabilities, and the class with the highest probability is predicted as a target. The architecture consists of convolutional, pooling, batch normalization, dropout and dense layers. The lower part of the architecture, enclosed inside the red square in Fig.~\ref{fig_HCRNet_architecture}, is similar to VGG16 architecture up to block4\_conv2 layer and acts as a feature extractor. It has four convolutional blocks: the first has two convolution layers with 64 filters followed by a max-pooling layer, the second block has two convolutional layers with 128 filters followed by max-pooling, the third block has three convolutional layers with 256 filters followed by max-pooling, and the last convolutional block has two convolutional layers with 512 filters. All the convolutional layers use a stride of one and padding as `same', and all the pooling layers use a stride of two and padding as `valid'. The convolutional blocks are followed by one batch-normalization layer and two dense layers each of which has 512 neurons and is followed by batch-normalization + dropout (with 35\% rate) layers. The last layer is a dense layer with neurons equal to the number of output classes. All convolutional and dense layers use ReLU as an activation function except the output dense layer which uses softmax because it is faster and helps to avoid the gradient vanishing problem. Categorical cross-entropy is used as a loss function with Root Mean Square Propagation (RMSprop)\footnote{\url{http://www.cs.toronto.edu/~tijmen/csc321/slides/lecture_slides_lec6.pdf}} as an optimizer to update weights/parameters of the network. The complexity of a deep CNN, and hence of HCR-Net is $\Omega$($n$), where $n$ is the number of pixels in an image.

For an input vector $x$, label vector $p$, predicted probability vector $\hat{p}$ and for $C$ classes, ReLU, softmax, categorical cross-entropy and RMSprop's weight update rules are given below \cite{prince2023understanding}.}
\begin{equation}
    ReLU(x) = \max \left(0, x \right),
\end{equation}
\begin{equation}
    softmax(x)_i = \frac{exp(x_i)}{\sum_{j}^{ }exp(x_j)},
\end{equation}
\begin{equation}
\label{eq_categorical_crossentropy}
    l(p,\hat{p}) = - \sum_{i=1}^{C} p_i \log(\hat{p}_i),
\end{equation}
\begin{equation}
\label{eq_rmsprop}
\begin{array}{l}
    v_t = \beta v_{t-1} + (1-\beta) g_t^2,\\\\
    W_{t+1} = W_t - \alpha_t V_t^{-1/2} g_t\\\\
\text{with } V_t^{-1/2} = \text{diag}(v_t + \epsilon),
\end{array}
\end{equation}
{\color{black}where $v_t$ is velocity term, $g_t = \nabla f(W_t, \xi_t)$, $\beta \in \left[0,1\right]$, $\alpha_t$ is learning rate (also called as step size), $W_t,\; \xi_t$ are model weights and randomness at step $t$, and $\epsilon$ is a very small number for numerical stability.} The different layers of HCR-Net are discussed below.

\begin{table}[htb!]
	\centering
	\caption{Summary of different layers and parameters of the proposed HCR-Net for an example with 10 classes}
	\label{tab_HCR-Net_layers}
	\begin{tabular}{llr}
		\hline
		\textbf{Layer (type)} & \textbf{Output Shape} & \textbf{\#Params} \\
		\hline
		Input layer & (None, 32, 32, 3) & 0 \\
		block1\_conv1 (Conv2D) & (None, 32, 32, 64) & 1792 \\
		block1\_conv2 (Conv2D) & (None, 32, 32, 64) & 36928 \\
		block1\_pool (MaxPooling2D) & (None, 16, 16, 64) & 0 \\
		block2\_conv1 (Conv2D) & (None, 16, 16, 128) & 73856 \\
		block2\_conv2 (Conv2D) & (None, 16, 16, 128) & 147584 \\
		block2\_pool (MaxPooling2D) & (None, 8, 8, 128) & 0 \\
		block3\_conv1 (Conv2D) & (None, 8, 8, 256) & 295168 \\
		block3\_conv2 (Conv2D) & (None, 8, 8, 256) & 590080 \\
		block3\_conv3 (Conv2D) & (None, 8, 8, 256) & 590080 \\
		block3\_pool (MaxPooling2D) & (None, 4, 4, 256) & 0 \\
		block4\_conv1 (Conv2D) & (None, 4, 4, 512) & 1180160 \\
		block4\_conv2 (Conv2D) & (None, 4, 4, 512) & 2359808 \\
		batch\_normalization & (None, 4, 4, 512) & 2048 \\
		flatten (Flatten) & (None, 8192) & 0 \\
		dense (Dense) & (None, 512) & 4194816 \\
		batch\_normalization\_1 & (None, 512) & 2048 \\
		dropout (Dropout) & (None, 512) & 0\\
		dense\_1 (Dense) & (None, 512) & 262656 \\
		batch\_normalization\_2 & (Batch (None, 512) & 2048 \\
		dropout\_1 (Dropout) & (None, 512) & 0 \\
		dense\_2 (Dense) & (None, 10) & 5130\\
		\hline
	\end{tabular}
\end{table}

\paragraph{Convolution layers} are the heart and soul of CNN and also give the network its name. It applies convolution operation which is a repeated application of a set of weights, called a filter, to the input image, and generates a feature map and helps in learning some specific feature during training. So, the use of multiple filters generates multiple feature maps, each learning some aspect of the image. Convolutions are very useful for learning spatial relationships in the input and reducing parameters by sharing weights. Let there are $l$ input feature maps of size $m*m$, convolutional filter size is $n*n$ with stride $s$, padding $p$, number of feature maps $k$ and output size $o*o$, then the number of parameters and output size in a convolutional layer is given below.
\begin{equation}
\begin{array}{l}
\text{\#Params} = (n*n*l+1)*k,\\
o = \lfloor \dfrac{m + 2p - n}{s} \rfloor + 1.
\end{array}
\end{equation}

\paragraph{Pooling layers (PLs)} are commonly inserted after successive convolutional layers. Its function is to down-sample the feature maps obtained from convolutional layers. So, it helps in reducing computations and the number of parameters, hence it avoids over-fitting and helps in achieving local translation invariance. PL uses filters, smaller than feature maps, on patches of feature maps and summarizes the information. The most commonly used pooling operations are max pooling and average pooling, which return the most activated feature and average feature, respectively. Let $m*m$ be the input size of one feature map, $n*n$ be the filter size with stride $s$ and output $o*o$, then the number of parameters and output size in a pooling layer is given below.
\begin{equation}
\begin{array}{l}
\text{\#Params} = 0,\\
o = \lfloor \dfrac{m - n}{s} \rfloor + 1.
\end{array}
\end{equation}

\paragraph{Batch-normalization layers}
It is a technique for training very deep neural networks that normalizes inputs of neural network layers coming from previous layers, and since this is done in batches so the name batch-normalization \cite{Ioffe2015batch}. It helps to stabilize the training of deep neural networks and get faster convergence. Chen et al. \cite{Chen2019} argued that a combination of batch-normalization and dropout outperforms the baselines and gives better training stability and faster convergence. So, we have used a combination of batch-normalization and dropout with dense layers in HCR-Net. {\color{black}Let $x_i$ is $i$-th $d$-dimensional input point, say an image, and $x^k_i$ refers to its $k$-th dimension, $B$ be a mini-batch of data points of size $b$, then mean and variance over $B$ is,}
\begin{equation}
    \begin{array}{l}
    \mu_B = \dfrac{1}{b} \sum_{i=1}^{b} x_i, \text{ and } \sigma_B^2 = \dfrac{1}{b} \sum_{i=1}^{b} \left( x_i - \mu_B \right)^2.
    \end{array}
\end{equation}
For $d$-dimensional input, each dimension is normalized as,
\begin{equation}
\begin{array}{l}
\hat{x}_i^{(k)} = \dfrac{x_i^{(k)} - \mu_B^{(k)}}{\sqrt{\sigma_B^{(k)^2}+ \epsilon}}, \text{where $k\in \left[1, d\right], i \in \left[1, m\right]$,}
\end{array}
\end{equation}
$\epsilon$ is an arbitrarily small constant added for numerical stability, and the transform is given below,
\begin{equation}
y_i^{(k)} = \gamma^{(k)} x_i^{(k)} + \beta^{(k)},
\end{equation}
{\color{black}where $y_i$ refers to output corresponding to input $x_i$, and $\gamma^{(k)}$ and $\beta^{(k)}$ are parameters learned during training. So, batch-normalization transform is $BN_{\gamma^{(k)} \beta^{(k)}}: x^{(k)}_{i=1,2,..,m} \rightarrow y^{(k)}_{i=1,2,..,m}$, with output equal to input and number of parameters equal to $2d$.}

\paragraph{Dropout layers} randomly and temporarily remove a given percentage of neurons from the hidden layers. This is one of the regularization techniques in deep learning and helps neural networks to avoid over-fitting because it helps neural networks remove dependency on any specific neuron. This is not a computationally expensive regularization technique as it does not require any specific implementation or new parameters. The output from the dropout layer is equal to the input of the layer. Mostly, it is used after a dense layer, but it can be used with other layers.

\paragraph{Dense layers}
This is the most commonly used layer in neural networks. The dense layer, also called the fully connected layer, contains a given number of neurons each of which is connected to all neurons in the previous layer. The output layer is mostly a dense layer which has neurons equal to the number of classes, representing different class probabilities. Let $n_i$ and $n_o$ be the number of neurons in input and output for a dense layer, then {\color{black} the following equation provides the computation of the number of parameters (\#Params) and output size in a Dense layer of a neural network.}
\begin{equation}
\begin{array}{l}
\text{\#Params} = n_i * n_o + n_o,\\
\text{Output size} = n_o.
\end{array}
\end{equation}

Table~\ref{tab_HCR-Net_layers} presents different layers of HCR-Net along with their types, outputs and number of parameters. In the first phase, layers up to block4\_conv2 are initialized from pre-trained VGG16 layers and are frozen, i.e., parameters/weights are not updated, and the rest of the layers are trained, so the number of trainable parameters in the first phase is 4,465,674 of total parameters 9,744,202. Moreover, for the second phase, all layer weights are updated so the number of trainable parameters is 9,741,130 while 3,072 are non-trainable. The non-trainable parameters belong to the batch-normalization layer because for each dimension batch-normalization maintains four parameters to keep track of the distributions, out of which two parameters are non-trainable (i.e., moving\_mean and moving\_variance).

\subsection{Transfer learning}
\label{subsec_transfer_learning}
In deep learning, transfer learning is a technique to transfer knowledge learned on one task to another related task. For example, use of a pre-trained deep learning network, like VGG16 which is trained on ImageNet, for HCR research. Thus, transfer learning enables the reuse of pre-trained models on a new but related problem. Transfer learning is very useful in faster training, mostly better results and learning on small datasets which otherwise will need large amounts of data for deep learning models \cite{Deore2020}. For transfer learning, either pre-trained models can be used, or a source model can be trained first where a large amount of data is available and then the source model can be reused on the target problem. In both cases, the entire model can be reused or part of it. As discussed in Section~\ref{subsec_literature_DL}, transfer learning has already been studied in HCR and has helped to get better performance, e.g., \cite{Pramanik2018,Guha2020,Deore2020}. 

As shown in Fig.~\ref{fig_HCRNet_architecture}, HCR-Net architecture partly utilizes VGG16 up to block4\_conv2 layer and initializes those layers with the pre-trained VGG16. Thus, our use of transfer learning is novel in HCR research to the best of our knowledge as it reuses pre-trained model partly, unlike existing research which uses entire pre-trained models \cite{Pramanik2018,Deore2020}. Moreover, this approach enables to use of transfer learning without using complex models. Details on the training of HCR-Net are provided in Subsection~\ref{subsubsec_two_phases_training}.

\subsection{Image augmentation}
\label{subsec_augmentation}
Image augmentation is a data augmentation technique which helps to artificially expand the training dataset by creating modified versions of the training images. For example, an input image can be modified by rotation, shear, translation, flip (vertical or horizontal), zoom and their hybrid combinations etc. This helps to apply deep learning techniques for problems with limited datasets which otherwise might not be possible to train on such a dataset. Image augmentation also helps a model to generalize well on the test dataset because of its training on different variants of training images. Some HCR studies have already used image augmentation and reported improvements in the performance of the model, e.g., \cite{Chowdhury2019,Deore2020,Sufian2020}.

\begin{figure}[htb!]
	\centering
	\includegraphics[width=0.6\linewidth]{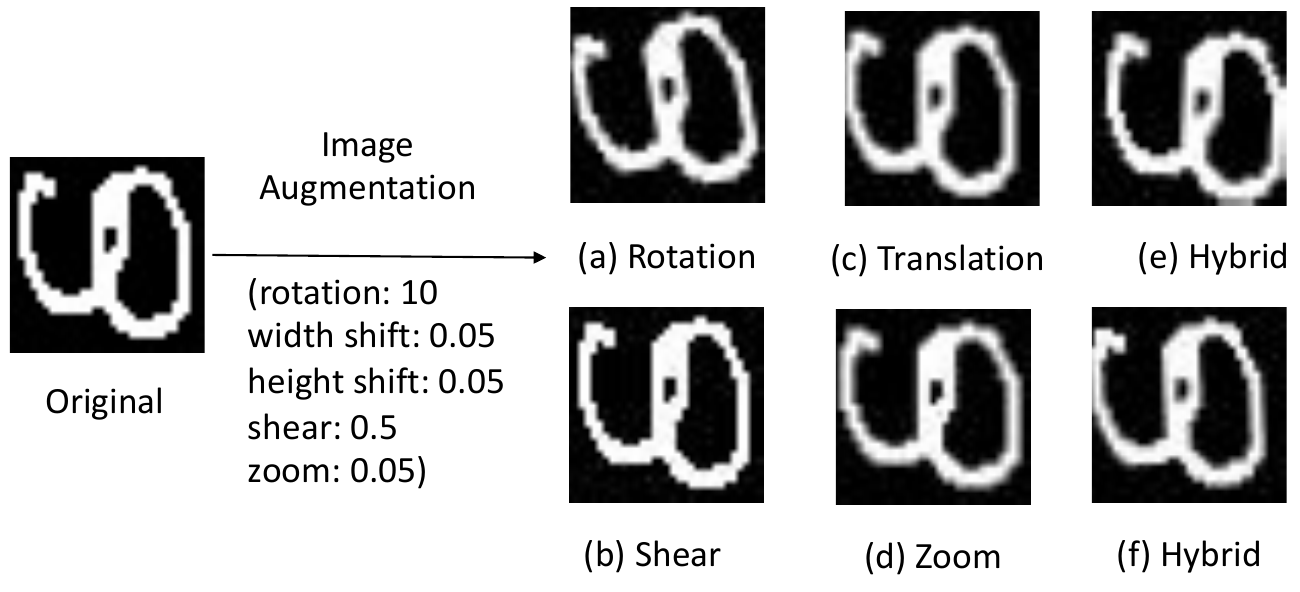}
	\caption{An example of mage augmentation using rotation, translation, shear, zoom and hybrid operations}
	\label{fig_HCRNet_image_augmentation}
\end{figure}

{\color{black}The application of image augmentation is crucial in improving model performance and generalization capabilities. Image augmentation encompasses a diverse array of transformative operations essential for enriching the training dataset. For example, rotation introduces variations in orientation by rotating images within a certain angle range (-45 to 45 degrees), offering distinct perspectives for model learning. Shearing involves selectively shifting parts of an image, such as slanting or skewing characters, enhancing the model's adaptability to different character structures. Translation displaces images horizontally or vertically, simulating changes in position and introducing spatial variations. Flipping, whether vertically or horizontally, mirrors the image and is approached cautiously in HCR to preserve character handedness. Zooming adjusts the image scale, providing the model with varying levels of detail, which is especially beneficial for capturing intricate character features. Hybrid combinations allow the simultaneous application of multiple operations, contributing to the creation of a more diverse set of augmented images. 
Fig.~\ref{fig_HCRNet_image_augmentation} presents an example of image augmentation on a character image and applies rotation, shear, translation, zoom and hybrid operations. It is to be noted that image augmentation should be applied carefully to HCR as it is different from other image classification tasks because it can distort the structure of a character and change its meaning, e.g., horizontal flip or large translation. For specific values for different image augmentation operations, please refer to Section~\ref{subsec_setting}.}

\subsection{Two phase training}
\label{subsubsec_two_phases_training}
HCR-Net trains in two phases due to transfer learning. In the first phase, parameters initialized from pre-trained weights of VGG16 are frozen and the rest of the parameters are updated. The model trains faster in the first phase than the second phase and is quite powerful because most of the time it can achieve up to 99\% of the final accuracy in just the first epoch and converges in a few epochs. The model obtained after the first phase is sufficient for most of the datasets like handwritten digit recognition.
\begin{table}[htb!]
	\centering
	\caption{Analysis of two phase training of HCR-Net}
	\label{tab_HCR-Net_phases}
	\begin{tabular}{ccccc}
		\hline
		\multirow{2}{*}{\textbf{Dataset}} & \multicolumn{2}{c}{\textbf{Without augmentation}} & \multicolumn{2}{c}{\textbf{With augmentation}} \\
		& \multicolumn{1}{c}{\textbf{\begin{tabular}[c]{@{}c@{}}First phase \\ first$\vert$Last epoch\end{tabular}}} & \multicolumn{1}{c}{\textbf{\begin{tabular}[c]{@{}c@{}}Second phase \\ last epoch\end{tabular}}} & \multicolumn{1}{c}{\textbf{\begin{tabular}[c]{@{}c@{}}First phase \\ first$\vert$Last epoch\end{tabular}}} & \multicolumn{1}{c}{\textbf{\begin{tabular}[c]{@{}c@{}}Second phase \\ last epoch\end{tabular}}} \\
		\hline
		Gurmukhi\_1.1 & 96.82$\vert$99.28 & 99.31 & 94.67$\vert$99.18 & 99.47 \\
		Kannada-mnist & 84.05$\vert$84.90 & 85.46 & 84.45$\vert$86.43 & 88.26 \\
		Mnist & 99.05$\vert$99.42 & 99.49 & 98.87$\vert$99.31 & 99.55 \\
		UCI Devanagari & 98.04$\vert$99.30 & 99.42 & 97.71$\vert$99.13 & 99.59\\
		\hline
	\end{tabular}
\end{table}

In the second training phase of HCR-Net, all the parameters of the network are updated. However, for the first few epochs learning rate is kept very small to avoid abrupt changes to the parameters and avoid losing information. Then, the learning rate is increased, as discussed in Subsec.~\ref{subsec_setting}. The second phase is useful for complex datasets, is computationally expensive and requires more epochs to converge with little improvements.

Table~\ref{tab_HCR-Net_phases} presents convergence analysis for two phases of HCR-Net. As it is clear from the table, without image augmentation, the first phase can get up to 99\% (and more in some cases) of final accuracy in just the first epoch. Moreover, there is a very slight improvement in the second phase. With image augmentation, test accuracy in the first epoch of first-phase training is lower as compared with without augmentation because training images have more diversity and hence more to learn due to modified variants of the images. Further, the with-augmentation second phase shows relatively more improvement over the first phase, as compared with without-augmentation.
\begin{figure}[htb!]
	\centering
	\begin{subfigure}[t]{0.48\textwidth}			
		\centering
		\includegraphics[width=\textwidth]{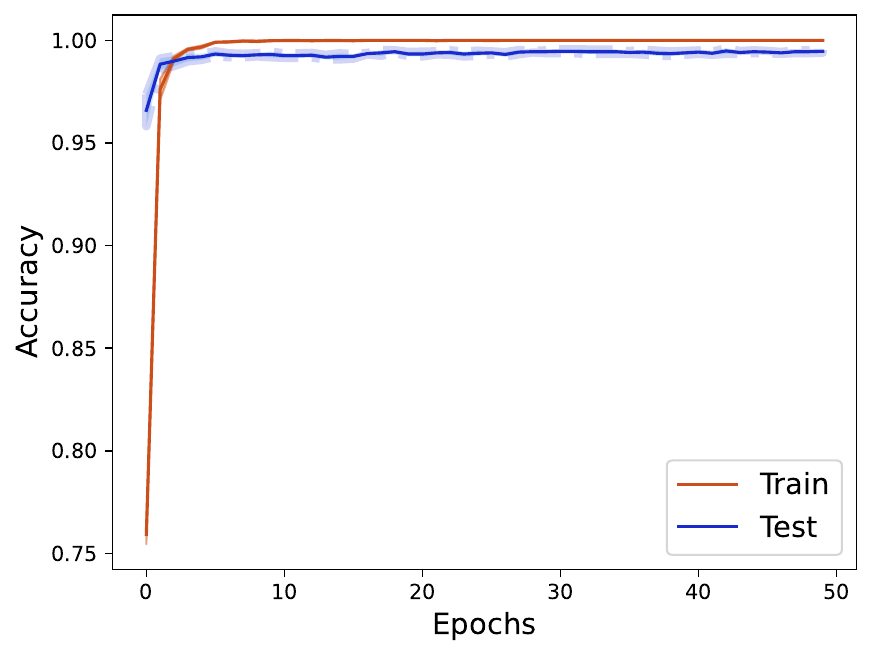}
		\caption{Accuracy against epochs}
		\label{subfig_machinistDS}
	\end{subfigure}%
	~ 
	\begin{subfigure}[t]{0.48\textwidth}			
		\centering
		\includegraphics[width=\textwidth]{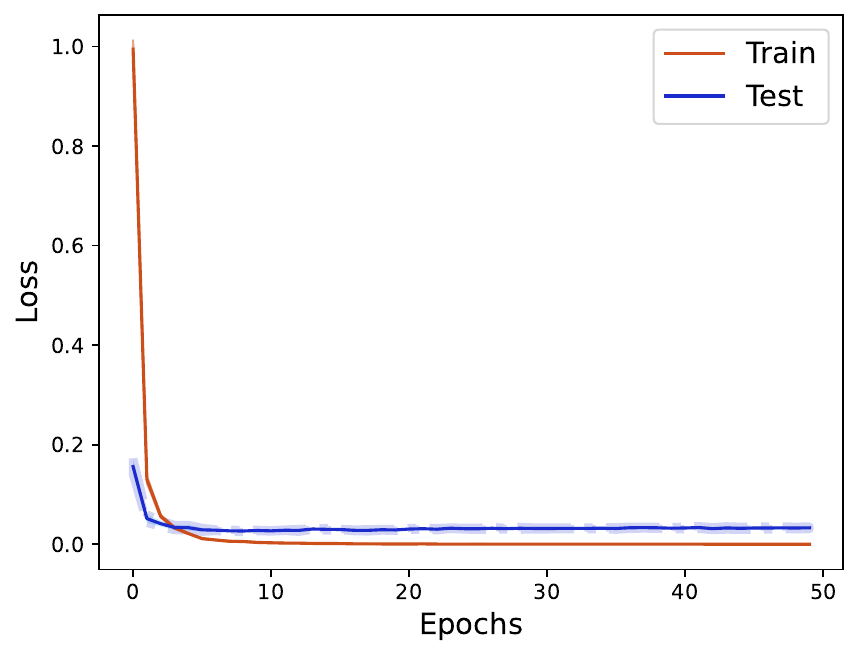}
		\caption{Loss against epochs}
		\label{subfig_forgerDS}
	\end{subfigure}	
	\caption{Convergence analysis of HCR-Net on Gurmukhi\_1.1 dataset, where the first phase takes the first 30 epochs and the second phase uses the remaining 20 epochs.}
	\label{fig_HCR-Net_convergence}
\end{figure}

Fig.~\ref{fig_HCR-Net_convergence} presents convergence analysis of two phases, averaged over five runs, of HCR-Net using RMSprop optimizer on Gurmukhi\_1.1 dataset, where the first phase runs for 30 epochs while the second phase runs for 20 epochs. The figure shows how accuracy improves and loss/error reduces with each epoch (i.e., pass through the data), where solid lines represent average results and shaded regions around the lines represent standard deviation in the performance. As it is clear from the figure, HCR-Net shows small deviations in the performance over multiple runs. HCR-Net also converges very quickly in a few epochs, and it is difficult to detect the slight performance improvement obtained in the second phase with bare eyes. Moreover, a small gap between test and train performance shows a nice convergence of the model without over-fitting, obtained with the use of dropout.

\section{Experimental results}
\label{sec_experiments}
This section presents statistics about datasets used in the experiments, provides experimental settings, compares HCR-Net against different benchmark studies and some of the state-of-the-art transfer learning techniques, and also discusses error analysis. 

\subsection{Datasets}
\label{subsec_datasets}
{\color{black}The experiments use publicly available datasets from Bangla, Punjabi, Hindi, English, Swedish, Urdu, Farsi, Tibetan, Kannada, Malayalam, Telugu, Marathi, Nepali and Arabic languages belonging to Bangla, Gurmukhi, Devanagari, Latin, Urdu, Farsi, Tibetan, Kannada, Malayalam, Telugu and Arabic scripts. The statistics of 41 datasets used in the experiments are given in Table~\ref{tab_datasets}. The datasets contain two online handwriting datasets (shown with `O' in the name), namely, IAPR TC-11 and UJIPenChars. The online datasets are first converted to offline images and then the proposed model is applied.}

\begin{longtable}{lrrrrrr}
\caption{Statistics for different HCR datasets}
\label{tab_datasets}
   \\ \hline
    \textbf{Dataset} & \multicolumn{1}{c}{\textbf{Writers}} & \multicolumn{1}{c}{\textbf{\begin{tabular}[c]{@{}l@{}}Samples\\ per class\end{tabular} }} & \multicolumn{1}{c}{\textbf{Classes}} & \multicolumn{1}{c}{\textbf{\begin{tabular}[c]{@{}l@{}}Training\\ Samples\end{tabular} }} & \multicolumn{1}{c}{\textbf{\begin{tabular}[c]{@{}l@{}}Testing\\ samples \end{tabular} }} & \multicolumn{1}{c}{\textbf{\begin{tabular}[c]{@{}l@{}}Total\\ samples\end{tabular} }} \\ \hline
    \textbf{Gurmukhi} & & & & & & \\
    HWRGurmukhi\_1.1 \cite{Jindal2019} & 1 & 100 & 35 & 2450 & 1050 & 3500 \\
    HWRGurmukhi\_1.2 \cite{Jindal2019} & 10 & 10 & 35 & 2450 & 1050 & 3500 \\
    HWRGurmukhi\_1.3 \cite{Jindal2019} & 100 & 1 & 35 & 2450 & 1050 & 3500 \\
    HWRGurmukhi\_2.1 \cite{Jindal2019} & 1 & 100 & 56 & 3920 & 1680 & 5600 \\
    HWRGurmukhi\_2.2 \cite{Jindal2019} & 10 & 10 & 56 & 3920 & 1680 & 5600 \\
    HWRGurmukhi\_2.3 \cite{Jindal2019} & 100 & 1 & 56 & 3920 & 1680 & 5600 \\
    HWRGurmukhi\_3.1 \cite{Jindal2019} & 200 & 1 & 35 & 4900 & 2100 & 7000 \\
    \hline 
    \textbf{Devanagari} & & & & & & \\
    Nepali (combined) \cite{Pant2012} & 40 & -- & 58 & -- & -- & 12,912 \\
    Nepali numeral \cite{Pant2012} & 40 & -- & 10 & -- & -- & 2880 \\
    Nepali vowels \cite{Pant2012} & 40 & -- & 12 & -- & -- & 2652 \\
    Nepali consonants \cite{Pant2012} & 40 & -- & 36 & -- & -- & 7380 \\
    \begin{tabular}[c]{@{}l@{}}Marathi numerals\end{tabular} \cite{Deore2020} & 100 & 100 & 10 & 800 & 200 & 1000 \\
    \begin{tabular}[c]{@{}l@{}}Marathi characters\end{tabular} \cite{Deore2020}& 100 & 100 & 48 & 3840 & 960 & 4800 \\
    \begin{tabular}[c]{@{}l@{}}Marathi combined\end{tabular} \cite{Deore2020} & 100 & 100 & 58 & 4640 & 1160 & 5800 \\
    \begin{tabular}[c]{@{}l@{}}UCI Devanagari\\ numerals\end{tabular} \cite{Acharya2015}& -- & 2000 & 10 & 17,000 & 3,000 & 20,000\\
    \begin{tabular}[c]{@{}l@{}}UCI Devanagari\\ characters\end{tabular} \cite{Acharya2015}& -- & 2000 & 36 & 61,200 &10,800 & 72,000 \\
    \begin{tabular}[c]{@{}l@{}}UCI Devanagari\\ total\end{tabular} \cite{Acharya2015} & -- & 2000 & 46 & 78,200 & 13,800 & 92,000\\
    \begin{tabular}[c]{@{}l@{}}CMATERdb\_3.2.1\\ Devanagari\\Numeral\end{tabular} \cite{Das2012,Das2012b} & -- & -- & 10 & 2400 & 600 & 3000 \\
    \begin{tabular}[c]{@{}l@{}}IAPR TC-11 (O) \cite{Santosh2010}\end{tabular} & 25 & -- & 36 & -- & -- & 1800 \\
    \hline
    \textbf{Bangla} & & & & & & \\
    \begin{tabular}[c]{@{}l@{}}CMATERdb\_3.1.1\\ (Bangla numeral) \end{tabular} \cite{Das2012}&-- &-- & 10 & 4089 & 1019 & 5108\\
    \begin{tabular}[c]{@{}l@{}}CMATERdb\_3.1.2\\ (Bangla character)\end{tabular} \cite{Das2015} & -- & 300 & 50 & 12,000 & 3,000 & 15,000 \\
    \begin{tabular}[c]{@{}l@{}}CMATERdb~3.1.3.3\\ (Bangla compound\\ character) \end{tabular} \cite{Das2014}& 335 & -- & 171 &44,152 & 11,126 & 55,278\\
    \begin{tabular}[c]{@{}l@{}}Banglalekha-isolated\\ numeral\end{tabular} \cite{Biswas2017} & -- & -- & 10 & 15802 & 3946 & 19748\\
    \begin{tabular}[c]{@{}l@{}}Banglalekha-isolated\\ character\end{tabular} \cite{Biswas2017} & -- & -- & 50 & 79179 & 19771 & 98950 \\
    \begin{tabular}[c]{@{}l@{}}Banglalekha-isolated\\ combined\end{tabular} \cite{Biswas2017} &-- & -- & 84 & 132914 & 33191 & 166105 \\
    ISI Bangla \cite{Bhattacharya2005} & -- & -- & 10 & 19392 & 4000 & 23,392 \\
    \hline
    \textbf{Latin} & & & & & & \\
    UJIPenchars (O) \cite{Llorens2008ujipenchars}& 11 & -- & 35 & 1240 & 124 & 1364 \\
    mnist \cite{Lecun1998} & -- & 7,000 & 10 & 60,000 & 10,000 & 70,000 \\
    ARDIS-II \cite{Kusetogullari2019}& -- & -- & 10& 6602& 1000 & 7602\\
    ARDIS-III \cite{Kusetogullari2019}&-- & -- & 10& 6600& 1000 & 7600\\
    ARDIS-IV \cite{Kusetogullari2019}& -- & -- & 10 & 6600& 1000 & 7600\\
    \hline 
    \textbf{Malayalam}* & & & & & & \\
    Amrita\_MalCharDb \cite{Manjusha2019} & 77 & -- & 85 & 17,236 & 6,360 & 29,302\\
    Malayalam\_DB \cite{Manjusha2018}& 77 & -- & 85 & 22,942 & 6,360 & 29,302\\
    \hline
    \textbf{Telugu} & & & & & & \\
    \begin{tabular}[c]{@{}l@{}}CMATERdb~3.4.1\\ (Telugu numeral) \end{tabular} \cite{Das2012}& -- & -- & 10 & 2400 & 600 & 3000 \\
    \hline
    \textbf{Kannada} & & & & & & \\
    Kannada-mnist \cite{Prabhu2019}& -- & -- & -- & 10 & 60,000 & 10,240 \\
    & & & & & 60,000 & 10,000 \\
    \hline
    \textbf{Urdu} & & & & & & \\
    Urdu \cite{Ali2020pioneer} & -- & -- & 10 & 6,606 & 1,414 & 8020 \\
    \hline
    \textbf{Farsi} & & & & & & \\
    Farsi \cite{Khosravi2007}& -- & -- & 10 & 60,000 & 20,000 & 80,000 \\
    \hline
    \textbf{Tibetan} & & & & & & \\
    Tibetan-mnist & -- & -- & 10 & 14,214 & 3,554 & 17,768\\
    \hline
    \textbf{Arabic} & & & & & & \\
    MADBase& -- & -- & 10 & 60,000 & 10,000 &70,000 \\
    \hline
    \multicolumn{4}{l}{* Both datasets are the same but differ in the splitting.}\\
    
\end{longtable}

\begin{table}[htb!]
	\setlength{\tabcolsep}{0.5pt}
	\renewcommand{\arraystretch}{3}
	\centering
	\caption{Few handwriting samples of some scripts}
	\label{tab_data_samples}
	\begin{tabular}{lllllllllll}
		Devanagari & \includegraphics[align=c,width=0.055\linewidth,height=0.055\linewidth]{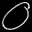} & \includegraphics[align=c,width=0.055\linewidth,height=0.055\linewidth]{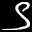}& \includegraphics[align=c,width=0.055\linewidth,height=0.055\linewidth]{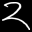} & \includegraphics[align=c,width=0.055\linewidth,height=0.055\linewidth]{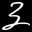}& \includegraphics[align=c,width=0.055\linewidth,height=0.055\linewidth]{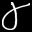}& \includegraphics[align=c,width=0.055\linewidth,height=0.055\linewidth]{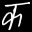}& \includegraphics[align=c,width=0.055\linewidth,height=0.055\linewidth]{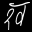}& \includegraphics[align=c,width=0.055\linewidth,height=0.055\linewidth]{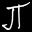}& \includegraphics[align=c,width=0.055\linewidth,height=0.055\linewidth]{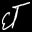}& \includegraphics[align=c,width=0.055\linewidth,height=0.055\linewidth]{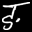} \\
		
		Bangla & \includegraphics[align=c,width=0.055\linewidth,height=0.055\linewidth]{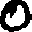} & \includegraphics[align=c,width=0.055\linewidth,height=0.055\linewidth]{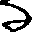}& \includegraphics[align=c,width=0.055\linewidth,height=0.055\linewidth]{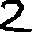} & \includegraphics[align=c,width=0.055\linewidth,height=0.055\linewidth]{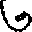}& \includegraphics[align=c,width=0.055\linewidth,height=0.055\linewidth]{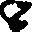}& \includegraphics[align=c,width=0.055\linewidth,height=0.055\linewidth]{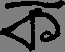}& \includegraphics[align=c,width=0.055\linewidth,height=0.055\linewidth]{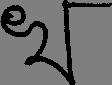}& \includegraphics[align=c,width=0.055\linewidth,height=0.055\linewidth]{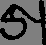}& \includegraphics[align=c,width=0.055\linewidth,height=0.055\linewidth]{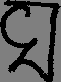}& \includegraphics[align=c,width=0.055\linewidth,height=0.055\linewidth]{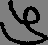} \\
		
		Latin & \includegraphics[align=c,width=0.055\linewidth,height=0.055\linewidth]{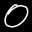} & \includegraphics[align=c,width=0.055\linewidth,height=0.055\linewidth]{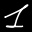}& \includegraphics[align=c,width=0.055\linewidth,height=0.055\linewidth]{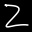} & \includegraphics[align=c,width=0.055\linewidth,height=0.055\linewidth]{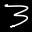}& \includegraphics[align=c,width=0.055\linewidth,height=0.055\linewidth]{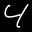}& \includegraphics[align=c,width=0.055\linewidth,height=0.055\linewidth]{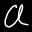}& \includegraphics[align=c,width=0.055\linewidth,height=0.055\linewidth]{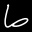}& \includegraphics[align=c,width=0.055\linewidth,height=0.055\linewidth]{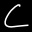}& \includegraphics[align=c,width=0.055\linewidth,height=0.055\linewidth]{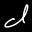}& \includegraphics[align=c,width=0.055\linewidth,height=0.055\linewidth]{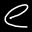} \\ 
	\end{tabular}
\end{table}

{\color{black}Few samples of some scripts are presented in Table~\ref{tab_data_samples}, where samples are taken from UCI Devanagari, CMATERdb~3.1.1 \& CMATERdb~3.1.2 (Bangla) and UJIPenChars (Latin) datasets. This highlights structural differences in different scripts, e.g., a horizontal line above the character is used in Devanagari and Bangla scripts but not in Latin script. This table also shows some noise (for Bangla) in recording different characters, which makes HCR a challenging task.}

\subsection{Experimental setup}
\label{subsec_setting}
Different hyperparameters of the HCR-Net, e.g., learning rate, mini-batch size, number of layers, neurons and optimizer are selected by trial over a range of values. Each experiment uses a mini-batch size of 32 and RMSprop as an optimizer. Although, RMSprop and Adam (Adaptive Moment Estimation), which are popular in HCR research, show similar test accuracy but RMSprop is selected because it takes less time for training. Image augmentation uses rotation of 10 degrees, horizontal and vertical shift of 0.05, shear of 0.5 and zoom of 0.05. Experiments use a staircase learning rate to better control the convergence of the learning algorithms. The first phase starts with a high learning rate of 1e-4 to get faster convergence up to five epochs, and then the learning rate is decreased to 5e-5 for the rest of the epochs.

In the second phase, up to five epochs learning rate is 1e-7 to avoid abrupt changes in weights, for the rest of the epochs except the last five, the learning rate is increased to 5e-6 and then further decreased to 1e-6 in the last five epochs. The number of epochs required to train HCR-Net is dependent on the dataset and trained until test accuracy becomes stable. Generally, without image augmentation, the first phase is run for 30 epochs and the second phase is run for at least 20 epochs. With image augmentation, the first phase is run for 10 epochs and the second phase is run for at least 50 epochs, this is because training with image augmentation learns more on diverse images than without augmentation and takes longer to converge. {\color{black}All the experiments are implemented using Keras library\footnote{\url{https://keras.io}}, averaged over five runs and executed on MacBook Pro (RAM 16 GB, Core-i7).}

The datasets are partitioned into 80:20 ratios for train and test, respectively wherever test sets are not separately available. {\color{black}The experimental results are reported on the test dataset using accuracy, precision, recall and F1-score. But, as it is clear from experiments with different datasets, all the metrics have almost similar values due to class-balanced datasets, so accuracy is a sufficient metric for HCR. Some authors also present test error/cost as a metric, since cost is dependent on the objective function/model used for recognition so cost is not a good metric and is not used. Moreover, test accuracy is reported at the last epoch of the network training, unlike some authors reporting the best accuracy which is not the correct way. This is because, during the training of a model, there may be a spike in test accuracy, i.e., the optimizer may enter the best solution region for a given test set but that might not give the best generalization at that epoch because the decision to select the model is based on the test dataset. So, the test accuracy is reported using test accuracy at the last epoch and best test accuracy during the training but only test accuracy at the last epoch is used to compare with the existing literature. We argue that either the results should be reported at the last epoch of training or reports should be reported on a separate dataset not used to decide the model, e.g., dividing the dataset into train, validation and test sets where train and validation may be used for training and final performance should be reported on the test set. Similarly, the training accuracy is also not a good metric for reporting HCR results because that does not reflect the generalization of the model and the model may be overfitting on the training dataset.}

\begin{table}[htb!]
		\centering
		\caption{Performance of HCR-Net on Gurmukhi script without$\vert$with augmentation}
		\label{tab_performance_Gurmukhi}
		\begin{tabular}{lrrrrr}
			\hline
			\multicolumn{1}{c}{\multirow{2}{*}{\textbf{Dataset}}} & \multicolumn{1}{c}{\multirow{2}{*}{\textbf{Precision}}} & \multicolumn{1}{c}{\multirow{2}{*}{\textbf{Recall}}} & \multicolumn{1}{c}{\multirow{2}{*}{\textbf{F1-score}}} & \multicolumn{2}{c}{\textbf{Accuracy}} \\
			\multicolumn{1}{c}{} & \multicolumn{1}{c}{} & \multicolumn{1}{c}{} & \multicolumn{1}{c}{} & \multicolumn{1}{c}{\textbf{At last epoch}} & \multicolumn{1}{c}{\textbf{Best}} \\
			\hline
		HWRGurmukhi\_1.1 & 99.34$\vert$99.48 & 99.31$\vert$99.47 & 99.31$\vert$99.47 & 99.31$\vert$99.47 & 99.49$\vert$99.52 \\
		HWRGurmukhi\_1.2 & 99.53$\vert$98.31 & 99.50$\vert$98.17 & 99.50$\vert$98.16 & 99.50$\vert$98.17 & 99.81$\vert$98.63 \\
		HWRGurmukhi\_1.3 & 96.71$\vert$96.02 & 96.53$\vert$95.64 & 96.53$\vert$95.66 & 96.53$\vert$95.64 & 96.69$\vert$95.96 \\
		HWRGurmukhi\_2.1 & 98.95$\vert$98.91 & 98.94$\vert$98.83 & 98.93$\vert$98.83 & 98.94$\vert$98.83 & 99.18$\vert$99.02 \\
		HWRGurmukhi\_2.2 & 94.58$\vert$93.85 & 94.11$\vert$93.37 & 93.95$\vert$93.20 & 94.11$\vert$93.37 & 94.30$\vert$93.76 \\
		HWRGurmukhi\_2.3 & 93.74$\vert$93.52 & 93.37$\vert$93.25 & 93.19$\vert$93.04 & 93.37$\vert$93.25 & 93.68$\vert$93.45 \\
		HWRGurmukhi\_3.1 & 97.40$\vert$96.90 & 97.34$\vert$96.68 & 97.33$\vert$96.66 & 97.34$\vert$96.68 & 97.49$\vert$96.99 \\
			\hline
		\end{tabular}
	\end{table}
	
\begin{table}[htb!]
		\centering
		\caption{Recognition rates on Gurmukhi datasets}
		\label{tab_Gurmukhi_results}
		\begin{tabular}{p{0.15\linewidth}p{0.1\linewidth}p{0.4\linewidth}p{0.08\linewidth}}
			\hline
			\textbf{Dataset} & \textbf{Reference} & \textbf{Methodology} & \multicolumn{1}{c}{\textbf{Accuracy}} \\
			\hline
			HWRGurmukhi\_1.1 & \cite{Jindal2019} & Random forest with diagonal features & 97.40 \\
			& HCR-Net & our work & \textbf{99.47} \\
			\hline
			HWRGurmukhi\_1.2 & \cite{Jindal2019} & Random forest with intersection and open end points features & 93.50 \\
			& HCR-Net & our work & \textbf{99.50} \\
			\hline
			HWRGurmukhi\_1.3 & \cite{Jindal2019} & MLP with directional features & 91.70 \\
			& HCR-Net & our work & \textbf{96.53} \\
			\hline
			HWRGurmukhi\_2.1 & \cite{Jindal2019} & Random forest with diagonal features & 92.60 \\
			& HCR-Net & our work & \textbf{98.94}\\
			\hline
			HWRGurmukhi\_2.2 & \cite{Jindal2019} & MLP with zoning features & 91.50 \\
			& HCR-Net & our work & \textbf{94.11} \\
			\hline
			HWRGurmukhi\_2.3 & \cite{Jindal2019} & Random forest with intersection and open end points features & 85.30 \\
			& HCR-Net & our work & \textbf{93.37}\\
			\hline
			HWRGurmukhi\_3.1 & \cite{Jindal2019} & Random forest with diagonal features & 90.50 \\
			& HCR-Net & our work & \textbf{97.34} \\
			\hline
		\end{tabular}
	\end{table}
	
\subsection{Preprocessing}
\label{subsec_preprocessing}
{\color{black}This paper does not use extensive preprocessing but simple preprocessing techniques as a generic architecture is developed for different scripts. HCR-Net uses greyscale character images of size $32\times32$ as inputs, and if image augmentation is turned on, then during the training on the fly applies image augmentation of rotation (10 degrees), translation (0.05), sheer (0.05), zoom (0.05) and hybrid combinations. Wherever possible, the character images are plotted against a black background to simplify the computations. All image pixel intensities are scaled to a range of 0 to 1.}

\subsection{Comparative study}
\label{subsec_comparision}
This subsection provides experimental results and comparisons with the literature. The performance of HCR-Net is reported on the test dataset using accuracy, precision, recall and F1-score without and with augmentation, respectively and is separated using `$\vert$'. We also present the best accuracy, in addition to accuracy at the last epoch just to show that the best test accuracy during training is almost all the time more than accuracy at the last epoch. However, we compare only test accuracy at the last epoch with the existing results. The following subsections discuss the results for different scripts.

\subsubsection{Gurmukhi script}
\label{subsubsec_Gurmukhi}
Table~\ref{tab_performance_Gurmukhi} presents the performance of HCR-Net on the Gurmukhi script. All the performance metrics for each dataset show similar results because datasets are class-balanced. {\color{black}HCR-Net performs very well even though the datasets are quite small, and that is because of the power of transfer learning from VGG16. Moreover, image augmentation shows improvement only on HWRGurmukhi\_1.1 but for the rest of the datasets it leads to a reduction in performance. This is because the datasets are collected in controlled environments and have lesser noise than real-world handwriting so expansion of the datasets with image augmentation techniques does not improve recognition.}
	
Table~\ref{tab_Gurmukhi_results} presents a comparative study of HCR-Net against the state-of-art results. Since these are recently released public datasets so there is not much literature to compare. From the table, it is clear that HCR-Net outperforms existing results and provides new benchmarks on all seven datasets, and shows up to seven percent improvement in the test accuracy. This is because \cite{Jindal2019} has used traditional machine learning techniques with handcrafted features. In the comparison of 1.1 with 1.3 or comparison of 2.1 with 2.3, cases with an equal number of samples and classes but 1 and 100 writers, respectively, show a decrease in test accuracy with an increase in number of writers. This makes the point that different people have different writing styles which impact the performance.

\subsubsection{Devanagari script}
\label{subsubsec_Devanagari}
\begin{table}[htb!]
\centering
\caption{Performance of HCR-Net on Devanagari script without$\vert$with augmentation}
\label{tab_performance_Devanagari}
\begin{tabular}{lrrrrr}
	\hline
	\multicolumn{1}{c}{\multirow{2}{*}{\textbf{Dataset}}} & \multicolumn{1}{c}{\multirow{2}{*}{\textbf{Precision}}} & \multicolumn{1}{c}{\multirow{2}{*}{\textbf{Recall}}} & \multicolumn{1}{c}{\multirow{2}{*}{\textbf{F1-score}}} & \multicolumn{2}{c}{\textbf{Accuracy}} \\
	\multicolumn{1}{c}{} & \multicolumn{1}{c}{} & \multicolumn{1}{c}{} & \multicolumn{1}{c}{} & \multicolumn{1}{c}{\textbf{At last epoch}} & \multicolumn{1}{c}{\textbf{Best}} \\
	\hline
	\begin{tabular}[c]{@{}l@{}}UCI Devanagari \\ numeral\end{tabular}& 99.97$\vert$99.99 & 99.97$\vert$99.99 & 99.97$\vert$99.99 & 99.97$\vert$99.99 & 99.99$\vert$100.00 \\
	\begin{tabular}[c]{@{}l@{}}UCI Devanagari \\ character\end{tabular} & 99.33$\vert$99.45 & 99.33$\vert$99.45 & 99.33$\vert$99.45 & 99.33$\vert$99.45 & 99.34$\vert$99.48 \\
	\begin{tabular}[c]{@{}l@{}}UCI Devanagari \\ combined\end{tabular} & 99.42$\vert$99.59 & 99.42$\vert$99.59 & 99.42$\vert$99.59 & 99.42$\vert$99.59 & 99.45$\vert$99.61 \\
	
	\begin{tabular}[c]{@{}l@{}}Marathi \\ numeral\end{tabular} 	& 99.05$\vert$98.86 & 99.00$\vert$98.80 & 99.00$\vert$98.80 & 99.00$\vert$98.80 & 99.20$\vert$99.30 \\
	\begin{tabular}[c]{@{}l@{}}Marathi \\ character\end{tabular} & 94.36$\vert$95.34 & 94.12$\vert$94.97 & 94.10$\vert$95.00 & 94.12$\vert$94.97 & 94.33$\vert$95.20 \\
	\begin{tabular}[c]{@{}l@{}}Marathi \\ combined\end{tabular} & 95.02$\vert$95.80 & 94.77$\vert$95.51 & 94.76$\vert$95.52 & 94.77$\vert$95.51 & 94.96$\vert$95.84 \\
	
	\begin{tabular}[c]{@{}l@{}}CMATERdb~3.2.1 \end{tabular} & 98.67$\vert$98.26 & 98.63$\vert$98.23 & 98.64$\vert$98.24 & 98.63$\vert$98.23 & 99.03$\vert$98.57 \\
	
	Nepali combined & 95.03$\vert$95.20 & 94.68$\vert$94.80 & 94.66$\vert$94.77 & 94.99$\vert$95.10 & 95.17$\vert$95.28 \\
	Nepali numeral & 99.79$\vert$99.59 & 99.79$\vert$99.58 & 99.79$\vert$99.58 & 99.79$\vert$99.58 & 99.89$\vert$99.82 \\
	Nepali vowel & 98.37$\vert$97.54 & 98.26$\vert$97.39 & 98.26$\vert$97.39 & 98.26$\vert$97.39 & 98.86$\vert$97.92 \\
	Nepali consonants & 93.98$\vert$94.07 & 93.56$\vert$93.60 & 93.50$\vert$93.53 & 93.56$\vert$93.60 & 93.96$\vert$94.09 \\
	
	IAPR TC-11 (O)	& 95.05$\vert$95.71 & 94.39$\vert$95.28 & 94.24$\vert$95.19 & 94.39$\vert$95.28 & 95.22$\vert$96.00 \\
	\hline
		\end{tabular}
	\end{table}
Table~\ref{tab_performance_Devanagari} presents the performance of HCR-Net on Nepali, Hindi and Marathi languages which share Devanagari script. All the performance metrics for each dataset show similar results because datasets are class-balanced. Image augmentation shows slight improvement on most of the datasets. {\color{black}It is to be noted that datasets with low performance, e.g., Marathi character and IAPR TC-11 (O) etc., show better improvements with image augmentation than others because others, like UCI Devanagari, have already reached a high level of performance and have a large number of samples.}

Table~\ref{tab_Devanagari_results} presents a comparative study of HCR-Net against the state-of-art results. From the table, it is clear that HCR-Net performs quite well and provides new benchmarks on Marathi numeral, UCI Devanagari (numeral and character) and Nepali (numeral, vowel consonants and combined). For the Nepali combined dataset, there is no reported result so there is not any literature to compare. {\color{black}The IAPR TC-11 dataset is a small online handwriting dataset which is converted to image form, and HCR-Net is able to beat the baseline model for the IAPR TC-11 dataset. So, this demonstrates HCR-Net's capability to recognize online handwriting, and it can perform better if datasets are larger. HCR-Net shows the largest improvements with the Nepali vowel dataset because the baseline uses handcrafted features with shallow learning, unlike HCR-Net which is powered by deep learning, image augmentation and transfer learning techniques. Additionally, the experiments were able to achieve a perfect score of 100\% test accuracy on the UCI numeral dataset four out of five times, which averaged 99.99\%. This is due to the large size of the UCI numeral dataset.}
		\begin{longtable}{p{0.2\linewidth}p{0.15\linewidth}p{0.5\linewidth}p{0.08\linewidth}}
			\caption{Recognition rates on Devanagari datasets}
		\label{tab_Devanagari_results}
            \\ \hline
			\textbf{Dataset} & \textbf{Reference} & \textbf{Methodology} & \multicolumn{1}{c}{\textbf{Accuracy}} \\ 
			\hline
	Marathi character & \cite{Deore2020} & fine-tuned VGG16 & \textbf{97.20} \\
	& HCR-Net & our work & 94.97 \\
	\hline
	Marathi numeral & \cite{Deore2020} & fine-tuned VGG16 & 95.50 \\
	& HCR-Net & our work & \textbf{99.00}\\
	\hline
	Marathi combined & \cite{Deore2020} & fine-tuned VGG16 & \textbf{96.55} \\
	& HCR-Net & our work & 95.51\\
	\hline
	IAPR TC-11 (O) & \cite{Santosh2010} & hierarchical stroke clustering and stroke-matching technique& 95.00 \\
	& \cite{Santosh2012stroke} & hierarchical stroke clustering and stroke-matching technique & \textbf{97.00}\\
	& HCR-Net & our work & 95.28\\
	\hline
	UCI Devanagari numeral & \cite{Acharya2015} & Deep CNN & 98.47 \\
	& & LeNet & 98.26 \\
	& 	\cite{Sarkar2012} & Back-propagation neural network with projection histogram& 92.20 \\
	& & Back-propagation neural network with chain code histogram& 92.70 \\
	& 	& Deep Auto-encoder network& 98.20 \\
	& \cite{Sarkhel2017} & a multi-column multi-scale CNN architecture + SVM & 99.50 \\
	& 	\cite{Deore2020} & fine-tuned VGG16 & 99.40 \\
	& 	HCR-Net & our work & \textbf{99.99}\\
	\hline
	UCI Devanagari character & \cite{Saha2020} & CNN & 93.00 \\
	& \cite{Gupta2019} & Multi-objective (recognition accuracy, redundancy of local regions and average recognition time per image) optimization to find the informative regions of the character image. Histogram of gradients (HOG) features+Convex hull features + quad-tree based Longest run features + SVM & 94.15 \\
	& \cite{Deore2020} & fine-tuned VGG16 & 97.80 \\
	& HCR-Net & our work & \textbf{99.45}\\
	\hline 
	UCI Devanagari combined & \cite{Acharya2015} & deep CNN architecture & 98.47 \\
	& \cite{Ram2018} & CNN architecture with 8 Layers & 96.90 \\
	& \cite{Mahapatra2020} & GAN + CNN Classifier & 97.38 \\
	& \cite{Guha2020} & deep CNN architecture & 99.54 \\
	& & DenseNet--121 & \textbf{99.60}* \\
	& HCR-Net & our work & 99.59\\
	\hline 
	CMATERdb~3.2.1 (Devanagari numeral) 
	& \cite{Ghosh2020} & Histogram of Oriented Pixel Positions and Point-Light Source-based Shadow with random forest & 98.01 \\
	& 	\cite{Guha2020} & deep CNN architecture & 98.70 \\
	& 	\cite{Das2012b} & modular Principal Component Analysis + quadtree based Longest-Run (QTLR) features + SVM & 98.70 \\
	& 	\cite{Sarkhel2017} & a multi-column multi-scale CNN architecture + SVM & \textbf{99.50} \\
	& 	HCR-Net & our work & 98.63\\
	\hline 
	Nepali numeral & \cite{Pant2012} & directional features, moment invariant features, Euler number, centroid of image, eccentricity and area of character skeleton + MLP and radial basis function & 94.44 \\
	& \cite{Ghosh2020} & Histogram of Oriented Pixel Positions and Point-Light Source-based Shadow with k-NN & 98.60\\
	& HCR-Net & our work & \textbf{99.79}\\
	Nepali vowel & \cite{Pant2012} & directional features, moment invariant features, Euler number, centroid of image, eccentricity and area of character skeleton + MLP and radial basis function & 86.04 \\
	& HCR-Net & our work & \textbf{98.26}\\
	Nepali consonants & \cite{Pant2012} & directional features, moment invariant features, Euler number, centroid of image, eccentricity and area of character skeleton + MLP and radial basis function & 80.25 \\
	& HCR-Net & our work & \textbf{93.60}\\
	Nepali (combined) & HCR-Net & our work & \textbf{95.10}\\
	\hline
	\multicolumn{4}{l}{* it appears authors have reported the max accuracy obtained during the training without averaging}\\
		\end{longtable}

\subsubsection{Latin script}
\label{subsubsec_Latin}
Table~\ref{tab_performance_Latin} presents the performance of HCR-Net on Swedish and English languages sharing Latin script. All the performance metrics for each dataset show similar results because the datasets are class-balanced. Image augmentation shows slight improvement on all of the datasets except ARDIS-III where there is a very slight drop in performance. {\color{black}This is because the datasets are large and already show good performance so there is little scope for improvement. The performance of UJIPenchars (O), an online handwriting dataset, is low compared to other datasets and shows the highest improvement with image augmentation. This is because it has only 1240 training points with 35 classes which are much smaller than the rest of the datasets.}
\begin{table}[htb!]
		\centering
		\caption{Performance of HCR-Net on Latin script without$\vert$with augmentation}
		\label{tab_performance_Latin}
		\begin{tabular}{lrrrrr}
			\hline
			\multicolumn{1}{c}{\multirow{2}{*}{\textbf{Dataset}}} & \multicolumn{1}{c}{\multirow{2}{*}{\textbf{Precision}}} & \multicolumn{1}{c}{\multirow{2}{*}{\textbf{Recall}}} & \multicolumn{1}{c}{\multirow{2}{*}{\textbf{F1-score}}} & \multicolumn{2}{c}{\textbf{Accuracy}} \\
			\multicolumn{1}{c}{} & \multicolumn{1}{c}{} & \multicolumn{1}{c}{} & \multicolumn{1}{c}{} & \multicolumn{1}{c}{\textbf{At last epoch}} & \multicolumn{1}{c}{\textbf{Best}} \\
			\hline
		mnist	& 99.50$\vert$99.55 & 99.48$\vert$99.54 & 99.49$\vert$99.55 & 99.49$\vert$99.55 & 99.54$\vert$99.61 \\
		ARDIS-II	& 99.68$\vert$99.78 & 99.68$\vert$99.78 & 99.68$\vert$99.78 & 99.68$\vert$99.78 & 99.80$\vert$99.96 \\
		ARDIS-III	& 99.84$\vert$99.80 & 99.84$\vert$99.80 & 99.84$\vert$99.80 & 99.84$\vert$99.80 & 100.00$\vert$99.96 \\
		ARDIS-IV	& 99.40$\vert$99.49 & 99.40$\vert$99.48 & 99.40$\vert$99.48 & 99.40$\vert$99.48 & 99.56$\vert$99.62 \\
		UJIPenchars (O)	& 85.70$\vert$86.05 & 86.86$\vert$87.71 & 84.76$\vert$85.20 & 87.74$\vert$88.23 & 89.19$\vert$89.68 \\
		\hline
		\end{tabular}
	\end{table}
	
Table~\ref{tab_Latin_results} presents a comparative study of HCR-Net against the state-of-art results. From the table, it is clear that HCR-Net performs quite well and provides new benchmarks on the ARDIS dataset (II, III and IV). mnist is a widely used benchmark dataset in computer vision and has extensive literature, here we have presented some representative studies only. {\color{black}Despite being a generic architecture, HCR-Net shows good performance on mnist with a very low error. ARDIS dataset is present in three different formats with different preprocessing and noise levels. HCR-Net outperforms on all datasets, including results given in literature without the name of the exact variant of ARDIS. Our proposed method shows large improvements as compared with the existing literature for ARDIS. Moreover, there is no result reported in the literature for the ARDIS-IV dataset so there is nothing to compare. It is further noted that HCR-Net performs consistently across different variants of the ARDIS dataset and shows near-perfect performance, despite different noise levels. This demonstrates the robust performance of the proposed script independent network.} HCR-Net does not perform well on the UJIPenchars (O), an online dataset, because it is a very small online handwriting dataset but the performance is still comparable to the existing literature.
\begin{table}[htb!]
		\centering
		\caption{Recognition rates on Latin datasets}
		\label{tab_Latin_results}
		\begin{tabular}{p{0.2\linewidth}p{0.15\linewidth}p{0.5\linewidth}p{0.08\linewidth}}
			\hline
			\textbf{Dataset} & \textbf{Reference} & \textbf{Methodology} & \multicolumn{1}{c}{\textbf{Accuracy}} \\ 
			\hline
	mnist & \cite{Manjusha2018} & CNN based on scattering transform-based wavelet filters & 99.31\\
	& \cite{Gupta2019} & Multi-objective (recognition accuracy, redundancy of local regions and average recognition time per image) optimization to find the informative regions of character image + SVM & 98.92 \\
	&	\cite{Wan2013} & CNN based architecture with DropConnect layer & \textbf{99.79}\\
	& HCR-Net & our work & 99.55\\
	\hline
	ARDIS & \cite{Kusetogullari2019} & CNN & 98.60\\
	&	\cite{Jiang2020} & pre-trained LeNet & 98.20 \\
	ARDIS-II & \cite{Ghosh2020} & Histogram of Oriented Pixel Positions and Point-Light Source-based Shadow with random forest & 94.27 \\
	& HCR-Net & our work & \textbf{99.78} \\
	ARDIS-III & \cite{Ghosh2020} & Histogram of Oriented Pixel Positions and Point-Light Source-based Shadow with sequential minimal optimization & 94.27 \\
	& HCR-Net & our work & \textbf{99.84} \\
	ARDIS-IV & HCR-Net & our work & \textbf{99.48} \\
	\hline
	UJIPenchars (O) & \cite{Llorens2008ujipenchars} & Microsoft Tablet PC SDK recognition engine & \textbf{91.60} \\
	& \cite{Prat2007two} & approximate dynamic time warping & 89.10 \\
	& HCR-Net & our work & 88.23 \\ 
	\hline
		\end{tabular}
	\end{table}

\subsubsection{Telugu, Malayalam and Kannada scripts}
\label{subsubsec_Telugu}
Table~\ref{tab_performance_south_indian} presents the performance of HCR-Net on Telugu, Malayalam and Kannada scripts' datasets. All the performance metrics for each dataset show similar results. Image augmentation shows slight improvement on most of the datasets except CMATERdb~3.4.1 where there is a very slight drop in performance. Kannada-mnist comes with two test sets, one of which is an out-of-distribution noisy test set, called Dig-mnist. Interestingly, image augmentation shows a sharp improvement of around three percent in the test accuracy for Dig-mnist, this is because image augmentation produces modified variants of images which are not present in the training set and helps in better generalization, which is more helpful in this case because Dig-mnist is an out-of-distribution noisy dataset. Moreover, best accuracy values are missing for the Kannada-mnist test set because it was used for final evaluation and Dig-mnist was used for evaluation during the training.

\begin{table}[htb!]
		\centering
		\caption{Performance of HCR-Net on Telugu, Malayalam and Kannada scripts}
		\label{tab_performance_south_indian}
		\begin{tabular}{lrrrrr}
			\hline
			\multicolumn{1}{c}{\multirow{2}{*}{\textbf{Dataset}}} & \multicolumn{1}{c}{\multirow{2}{*}{\textbf{Precision}}} & \multicolumn{1}{c}{\multirow{2}{*}{\textbf{Recall}}} & \multicolumn{1}{c}{\multirow{2}{*}{\textbf{F1-score}}} & \multicolumn{2}{c}{\textbf{Accuracy}} \\
			\multicolumn{1}{c}{} & \multicolumn{1}{c}{} & \multicolumn{1}{c}{} & \multicolumn{1}{c}{} & \multicolumn{1}{c}{\textbf{At last epoch}} & \multicolumn{1}{c}{\textbf{Best}} \\
			\hline
		\begin{tabular}[c]{@{}l@{}}CMATERdb \\ 3.4.1 \end{tabular} & 98.80$\vert$98.40 & 98.77$\vert$98.30 & 98.77$\vert$98.30 & 98.77$\vert$98.30 & 99.13$\vert$98.57 \\
		\hline
		Amrita\_MalCharDb	& 94.98$\vert$95.32 & 94.73$\vert$94.99 & 94.77$\vert$95.05 & 94.84$\vert$95.15 & 94.87$\vert$95.28 \\
		Malayalam\_DB & 95.12$\vert$95.45 & 94.87$\vert$95.16 & 94.91$\vert$95.20 & 94.95$\vert$95.32 & 95.02$\vert$95.42 \\
		\hline
		\begin{tabular}[c]{@{}l@{}}Kannada-mnist \\ (test-set)\end{tabular} & 98.08$\vert$98.30 & 98.05$\vert$98.27 & 98.05$\vert$98.26 & 98.05$\vert$98.27 & ------$\vert$------ \\
		\begin{tabular}[c]{@{}l@{}}Kannada-mnist \\ (Dig-mnist)	\end{tabular} & 86.00$\vert$88.57 & 85.46$\vert$88.26 & 85.45$\vert$88.28 & 85.46$\vert$88.26 & 86.61$\vert$88.72 \\
		\hline
		\end{tabular}
	\end{table}

Table~\ref{tab_south_indian_results} presents a comparative study of HCR-Net against the state-of-art results. From the table, it is clear that HCR-Net performs quite well and provides new benchmarks on Amrita\_MalCharDb, Malayalam\_DB and Kannada-mnist (Dig-mnist) datasets. CMATERdb 3.4.1 has few versions and it appears that \cite{Das2012b} used a different version as the dataset statistics are different than the one used in our work. {\color{black}We obtained a huge improvement of over 11\% on Dig-mnist, which is an out-of-distribution noisy dataset collected from a practical situation, because of image augmentation (as reported in Table~\ref{tab_performance_south_indian}) and transfer learning. This demonstrates the robustness of the proposed HCR-Net and its suitability for practical real-world applications where data are noisy and have different styles.}

\begin{table}[htb!]
		\centering
		\caption{Recognition rates on Telugu, Malayalam and Kannada datasets}
		\label{tab_south_indian_results}
		\begin{tabular}{p{0.2\linewidth}p{0.15\linewidth}p{0.5\linewidth}p{0.08\linewidth}}
			\hline
			\textbf{Dataset} & \textbf{Reference} & \textbf{Methodology} & \multicolumn{1}{c}{\textbf{Accuracy}} \\ 
			\hline
		CMATERdb~3.4.1 & 	\cite{Das2012b} & Modular Principal Component Analysis + quadtree based Longest-Run (QTLR) features + SVM & 99.20 \\
		& \cite{Sarkhel2017} & a multi-column multi-scale CNN architecture + SVM & \textbf{99.50}* \\
		& \cite{Ghosh2020} & Histogram of Oriented Pixel Positions and Point-Light Source-based Shadow with random forest& 99.03\\
		& HCR-Net & our work & 98.77 \\
		\hline
		Malayalam\_DB & \cite{Manjusha2018} & CNN based on scattering transform-based wavelet filters & 93.77\\
		& & SmallResnet based on scattering transform-based wavelet filters & 92.85 \\
		& & SmallResnet based on scattering transform-based wavelet filters + image augmentation & 95.27 \\
		& HCR-Net & our work & \textbf{95.32} \\
		Amrita\_MalCharDb & \cite{Manjusha2019} & CNN based on scattering transform-based wavelet filters as feature extractor and Linear SVM as classifier & 91.05 \\
		& HCR-Net & our work & \textbf{95.15} \\
		\hline
		Kannada-mnist (test set) & \cite{Prabhu2019} & End-to-end training using CNN based architecture & 96.80\\
		& \cite{Mahapatra2020} & GAN + CNN Classifier &  \textbf{98.70}\\
		& \cite{Saini2021} & deep residual network ResNeXt & 97.36\\
		& HCR-Net & our work & 98.27\\
		Kannada-mnist (Dig-mnist) & \cite{Prabhu2019} & End-to-end training using CNN based architecture & 76.10\\
		& \cite{Saini2021} & deep residual network ResNeXt & 79.06\\
		& HCR-Net & our work & \textbf{88.26}\\
		\hline
			\multicolumn{4}{l}{* authors used different training dataset}\\
		\end{tabular}
	\end{table}

\subsubsection{Bangla script}
\label{subsubsec_Bangla}
Bangla is one of the widely studied Indian scripts and it has several public datasets which further enhance the research of this script (refer to \cite{singh2018comprehensive} for a survey on Bangla handwritten numeral recognition). Table~\ref{tab_performance_Bangla} presents the performance of HCR-Net on Bangla script datasets. All the performance metrics for each dataset show similar results. {\color{black}Image augmentation shows slight improvement on most of the datasets except 3.1.1 and 3.1.2 where there is a slight drop in performance and 3.1.3.3 shows large improvements. This is because the 3.1.3.3 dataset has low performance compared to the others and hence there is more scope for improvement.}

\begin{table}[htb!]
		\centering
		\caption{Performance of HCR-Net on Bangla script without$\vert$with augmentation}
		\label{tab_performance_Bangla}
		\begin{tabular}{lrrrrr}
			\hline
			\multicolumn{1}{c}{\multirow{2}{*}{\textbf{Dataset}}} & \multicolumn{1}{c}{\multirow{2}{*}{\textbf{Precision}}} & \multicolumn{1}{c}{\multirow{2}{*}{\textbf{Recall}}} & \multicolumn{1}{c}{\multirow{2}{*}{\textbf{F1-score}}} & \multicolumn{2}{c}{\textbf{Accuracy}} \\
			\multicolumn{1}{c}{} & \multicolumn{1}{c}{} & \multicolumn{1}{c}{} & \multicolumn{1}{c}{} & \multicolumn{1}{c}{\textbf{At last epoch}} & \multicolumn{1}{c}{\textbf{Best}} \\
			\hline
		CMATERdb~3.1.1	& 98.86$\vert$98.52 & 98.84$\vert$98.49 & 98.84$\vert$98.49 & 98.84$\vert$98.49 & 98.98$\vert$98.74 \\
		CMATERdb~3.1.2 & 97.46$\vert$96.69 & 97.42$\vert$96.61 & 97.42$\vert$96.60 & 97.42$\vert$96.61 & 97.59$\vert$96.77 \\
		CMATERdb~3.1.3.3 & 89.09$\vert$92.53 & 88.75$\vert$92.17 & 88.81$\vert$92.21 & 88.73$\vert$92.19 & 88.81$\vert$92.35 \\
		ISI Bangla & 99.26$\vert$99.44 & 99.26$\vert$99.44 & 99.25$\vert$99.43 & 99.26$\vert$99.44 & 99.33$\vert$99.51 \\
		Bangalalekha\_isolated& 99.24$\vert$99.40 & 99.24$\vert$99.40 & 99.24$\vert$99.40 & 99.24$\vert$99.40 & 99.30$\vert$99.50 \\
		numerals&&&&&\\ 
		Bangalalekha\_isolated& 96.96$\vert$97.32 & 96.94$\vert$97.30 & 96.94$\vert$97.30 & 96.95$\vert$97.30 & 96.99$\vert$97.31 \\
		characters&&&&&\\
		Bangalalekha\_isolated& 95.74$\vert$96.21 & 95.69$\vert$96.18 & 95.69$\vert$96.18 & 95.70$\vert$96.19 & 95.74$\vert$96.20 \\
		combined&&&&&\\
		\hline
		\end{tabular}
\end{table}

\begin{table}[htb!]
        \centering
        \caption{Recognition rates on Bangla datasets}
        \label{tab_Bangla_results}
        \begin{tabular}{p{0.2\linewidth}p{0.15\linewidth}p{0.5\linewidth}p{0.08\linewidth}}
            \hline
            \textbf{Dataset} & \textbf{Reference} & \textbf{Methodology} & \multicolumn{1}{c}{\textbf{Accuracy}} \\ 
            \hline 
    CMATERdb~3.1.1 (Bangla numeral) & \cite{Das2012} & SVM classifier using GA for region subsampling of local features & 97.70 \\
    & \cite{Das2012b} & Modular Principal Component Analysis and Quad-tree based hierarchically derived Longest-Run features + SVM & 98.55 \\
    & \cite{Roy2014} & Axiomatic Fuzzy Set theory to calculate features’ combined class separability + quad-tree based longest-run feature set and gradient-based directional feature set + SVM & 97.45 \\
    & \cite{Sarkhel2017} & a multi-column multi-scale CNN architecture + SVM & \textbf{100.00}* \\
    & \cite{Keserwani2019} & spatial pyramid pooling and fusion of features from different layers of CNN& 98.80 \\
    & \cite{Deore2020} & fine-tuned VGG16 & 97.45 \\
    & \cite{Ghosh2020} & Histogram of Oriented Pixel Positions and Point-Light Source-based Shadow with random forest& 98.50 \\
    & HCR-Net & our work & 98.84 \\
    \hline
    CMATERdb~3.1.2 (Bangla basic character) 
    & \cite{Bhattacharya2006} & local chain code histograms + SVM & 92.14 \\
    & \cite{Sarkhel2017} & a multi-column multi-scale CNN architecture + SVM & \textbf{100.00} \\
    & \cite{Chowdhury2019} & CNN based architecture & 93.37 \\
    & \cite{Keserwani2019} & spatial pyramid pooling and fusion of features from different layers of CNN& 98.56 \\
    & \cite{Deore2020} & fine-tuned VGG16 & 95.83 \\
    & HCR-Net & our work & 97.42\\
    \hline 
    CMATERdb~3.1.3.3 (compound character) 
    & \cite{Das2015b} & Genetic algorithm based Two pass approach + SVM& 87.50 \\
    & \cite{Das2014} & A convex hull and quad tree-based features + SVM & 79.35\\
    & \cite{Roy2017} & & 90.33 \\
    & \cite{Sarkhel2017} & a multi-column multi-scale CNN architecture + SVM & \textbf{98.12}* \\
    & \cite{Pramanik2018b} & & 88.74 \\
    & \cite{Keserwani2019} & spatial pyramid pooling and fusion of features from different layers of CNN & 95.70 \\
    & HCR-Net & our work & 92.19\\
    \hline
    ISI Bangla& \cite{Jiang2020} &pre-trained LeNet  & 97.05 \\ 
    & \cite{Bhattacharya2009} & multilayer perceptron classifiers using wavelet-based multi-resolution features& 98.20 \\
    & \cite{Sarkhel2016} & Multi-objective (recognition accuracy and recognition cost per image) optimization to find the informative regions of character image + SVM & 98.23 \\
    & \cite{Gupta2019} & Multi-objective (recognition accuracy, redundancy of local regions and average recognition time per image) optimization to find the informative regions of character image + SVM & 98.61 \\
    & \cite{Sufian2020} & densely connected CNN and image augmentation& \textbf{99.78} \\ 
    & HCR-Net & our work & 99.44 \\
    \hline
    \begin{tabular}[c]{@{}l@{}}Banglalekha-isolated\\ combined\end{tabular} & \cite{Chowdhury2019} & CNN based architecture (2 Conv2D) & 95.25 \\
    & HCR-Net & our work & \textbf{96.19} \\
    \begin{tabular}[c]{@{}l@{}}Banglalekha-isolated\\ numbers\end{tabular}	& \cite{Jiang2020} & pre-trained LeNet & 94.86\\
    & HCR-Net & our work & \textbf{99.40} \\
    \begin{tabular}[c]{@{}l@{}}Banglalekha-isolated\\ characters\end{tabular}& HCR-Net & our work & \textbf{97.30} \\
    \hline
            \multicolumn{4}{l}{* authors used a different version of dataset}\\
        \end{tabular}
\end{table}

Table~\ref{tab_Bangla_results} presents a comparative study of HCR-Net against the state-of-art results. From the table, it is clear that HCR-Net performs quite well and provides a few new benchmarks on Banglalekha-isolated (numerals, characters and combined). It is observed that HCR-Net lags for complex problems with a large number of classes, like CMATERdb~3.1.3.3 which has 171 classes, and also trains slowly and takes a large number of epochs, 180 in this case. {\color{black}However, HCR-Net performs second-best for CMATERdb~3.1.3.3, while the rest show a large performance gap. It is also noted that among Bangla datasets, CMATERdb~3.1.3.3 obtains the lowest performance due large number of classes. Moreover, it is observed that a multi-column multi-scale CNN architecture proposed by \cite{Sarkhel2017} performs exceptionally well for the Bangla script.}

\subsubsection{Few other scripts}
\label{subsubsec_others}
Table~\ref{tab_performance_ohters} presents the performance of HCR-Net on Farsi, Urdu, Tibetan and Arabic scripts' datasets. All the performance metrics for each dataset show similar results. {\color{black}Image augmentation does not show consistent improvements on most of the datasets as there are slight changes in the performance which could be because of randomness associated with the experiments and the fact that these datasets already show high performance without image augmentation, leaving little space for further improvements.}

\begin{table}[htb!]
    \centering
    \caption{Performance of HCR-Net on Farsi, Urdu, Tibetan and Arabic scripts without$\vert$with augmentation}
    \label{tab_performance_ohters}
    \begin{tabular}{lrrrrr}
        \hline
        \multicolumn{1}{c}{\multirow{2}{*}{\textbf{Dataset}}} & \multicolumn{1}{c}{\multirow{2}{*}{\textbf{Precision}}} & \multicolumn{1}{c}{\multirow{2}{*}{\textbf{Recall}}} & \multicolumn{1}{c}{\multirow{2}{*}{\textbf{F1-score}}} & \multicolumn{2}{c}{\textbf{Accuracy}} \\
        \multicolumn{1}{c}{} & \multicolumn{1}{c}{} & \multicolumn{1}{c}{} & \multicolumn{1}{c}{} & \multicolumn{1}{c}{\textbf{At last epoch}} & \multicolumn{1}{c}{\textbf{Best}} \\
        \hline
    Farsi & 99.38$\vert$99.29 & 99.38$\vert$99.28 & 99.38$\vert$99.28 & 99.38$\vert$99.28 & 99.39$\vert$99.35 \\
    Urdu & 98.63$\vert$98.83 & 98.66$\vert$98.83 & 98.64$\vert$98.83 & 98.66$\vert$98.84 & 98.95$\vert$99.01 \\
    Tibetan-mnist & 99.37$\vert$99.26 & 99.37$\vert$99.27 & 99.37$\vert$99.26 & 99.35$\vert$99.24 & 99.43$\vert$99.41 \\
    MADBase & 99.21$\vert$99.26 & 99.21$\vert$99.25 & 99.21$\vert$99.25 & 99.21$\vert$99.25 & 99.32$\vert$99.33 \\
    \hline
    \end{tabular}
\end{table}

Table~\ref{tab_others_results} presents a comparative study of HCR-Net against the state-of-art results. From the table, it is clear that HCR-Net performs quite well and provides new benchmarks on Urdu, Tibetan-mnist and MADBase datasets. For FARSI, DenseNet based model performs the best followed by HCR-Net with a very small margin. {\color{black}It is also noted that among these scripts Urdu has the smallest dataset and HCR-Net performs the best with more than one and a half percent improvement over the baselines leading to near-perfect performance.}

\begin{table}[htb!]
    \centering
    \caption{Recognition rates Farsi, Urdu, Tibetan and Arabic scripts' datasets}
    \label{tab_others_results}
    \begin{tabular}{p{0.15\linewidth}p{0.13\linewidth}p{0.4\linewidth}p{0.08\linewidth}}
        \hline
        \textbf{Dataset} & \textbf{Reference} & \textbf{Methodology} & \multicolumn{1}{c}{\textbf{Accuracy}} \\ 
        \hline 
    FARSI& \cite{Parseh2017new} & PCA+SVM & 99.07\\
    & \cite{Akhlaghi2020farsi} & CNN based architecture & 99.34 \\
    & \cite{Bonyani2021persian} & DenseNet + data augmentation & \textbf{99.49}\\
    & HCR-Net & our work & 99.38 \\
    \hline
    Urdu& \cite{Jiang2020} & pre-trained LeNet & 97.31\\
    & \cite{Akhlaghi2020farsi} & CNN based architecture & 96.57 \\
    & \cite{Ali2020pioneer} & autoencoder and CNN architecture & 97.00\\
    & HCR-Net & our work & \textbf{98.84} \\
    \hline
    Tibetan-mnist& \cite{Jiang2020} & pre-trained LeNet & 98.31 \\
    & HCR-Net & our work & \textbf{99.35} \\
    \hline
    MADBase & \cite{Jiang2020} & pre-trained LeNet & 98.93 \\
    & \cite{Alkhawaldeh2021arabic} & LeNet + LSTM & 98.47\\
    & HCR-Net & our work & \textbf{99.25} \\
    \hline
    \end{tabular}
\end{table}

{\color{black}Thus, from these experiments, we conclude that HCR-Net is a script independent architecture which can handle different scripts. It performs very well and establishes several new benchmarks. It is observed that HCR-Net shows very high performance on datasets with a smaller number of classes, like numerals. The image augmentation component of HCR-Net shows great performance improvement when the test dataset is out-of-distribution and noisy, e.g., Dig-mnist (Table~\ref{tab_performance_south_indian}) and ARDIS (Table~\ref{tab_performance_Latin}. Transfer learning helps HCR-Net to get faster convergence, robust results and better generalization (Fig.~\ref{fig_HCR-Net_convergence} and Table~\ref{tab_HCR-Net_phases}).}

\subsubsection{Comparison with transfer learning techniques}
\label{subsubsec_sota}
\begin{figure}[htb!]
    \centering
    \begin{subfigure}[t]{0.48\textwidth}			
        \centering
        \includegraphics[width=\textwidth]{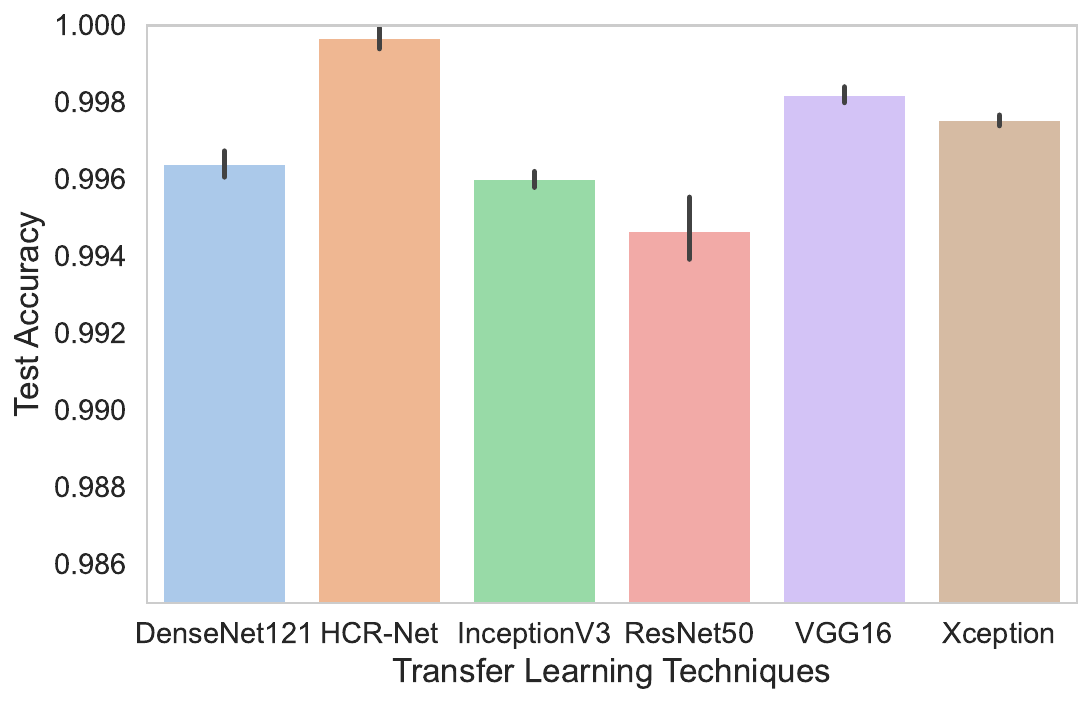}
        \caption{Comparison in terms of test accuracy}
        \label{subfig_sota_acc}
    \end{subfigure}	
    ~
    \begin{subfigure}[t]{0.48\textwidth}			
        \centering
        \includegraphics[width=\textwidth]{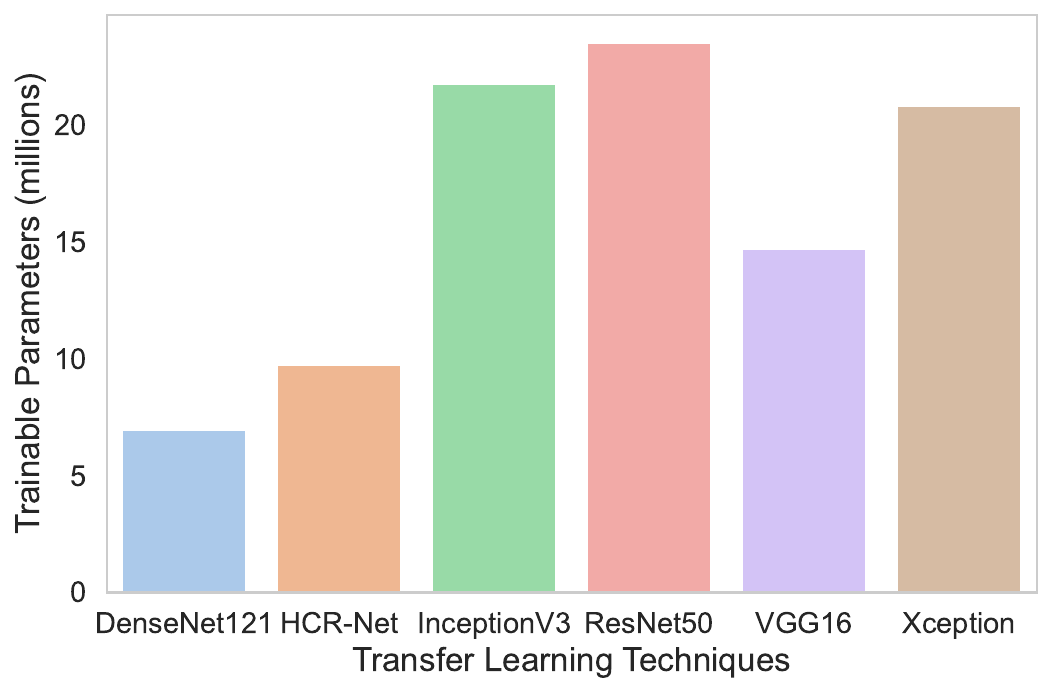}
        \caption{Comparison in terms of trainable parameters}
        \label{subfig_sota_params}
    \end{subfigure}%
    \caption{Performance comparison of HCR-Net against state-of-the-art transfer learning techniques for HCR}
	\label{fig_sota}
\end{figure}

{\color{black}Here, we compare HCR-Net against some of the popularly used state-of-the-art transfer learning techniques for HCR as VGG16 \cite{VGG16}, Xception \cite{chollet2017xception}, ResNet50 \cite{he2016deep}, InceptionV3 \cite{szegedy2016rethinking} and DenseNet121 \cite{huang2017densely}. Fig.~\ref{fig_sota} presents the comparative study in terms of test accuracy and number of trainable parameters to measure the computational performance using the UCI Devanagari Numeral dataset. The experimental setup for HCR-Net and other transfer learning techniques is the same. All methods train in two phases, where the first phase trains only the classifier layer while the second phase trains the entire network. From the figure, it is clear that HCR-Net significantly (p-value=0.00099 using Student's t-test) outperforms the rest of the transfer learning techniques, and VGG16 is the second-best technique. ResNet50 is the worst performer in terms of test accuracy. However, all the transfer learning approaches show impressive performance and that is why they are widely used in the HCR research. It is also observed that only HCR-Net shows fast convergence and could achieve high performance immediately after the first epoch (please refer to Subsec.~\ref{subsubsec_two_phases_training} for HCR-Net convergence, however, the convergence of the rest of the techniques is not shown here). In terms of computational efficiency, DenseNet121 and HCR-Net are the first and second best techniques, respectively, and have a large gap from the rest of the techniques. HCR-Net reduces the number of trainable parameters of the corresponding VGG16 by 34\% and thus is a computationally efficient technique.}

\subsection{Misclassification analysis}
\label{subsec_miss_classification_analysis}
In this subsection, the causes of misclassifications are analysed by taking examples of the UCI Devanagari numeral dataset and the IAPR-11 Devanagari dataset, where HCR-Net outperforms and lacks, respectively.

\begin{figure}[htb!]
	\centering
	\begin{subfigure}[t]{0.4\textwidth}		
		\includegraphics[width=\textwidth]{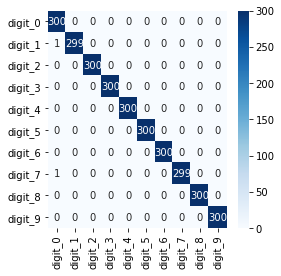}
		\caption{Confusion matrix of UCI Devanagari numeral dataset, showing only two misclassifications.}
		\label{subfig_miss_classification_confusion_mat}
	\end{subfigure}%

	\begin{subfigure}[t]{0.12\textwidth}			
		\centering
		\includegraphics[width=0.6\textwidth]{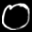}
		\caption{`digit\_0'}
		\label{subfig_miss_classification_1}
	\end{subfigure}%
	~
	\begin{subfigure}[t]{0.12\textwidth}			
		\centering
		\includegraphics[width=0.6\textwidth]{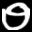}
		\caption{`digit\_7'}
		\label{subfig_miss_classification_2}
	\end{subfigure}%
	~ 
	\begin{subfigure}[t]{0.12\textwidth}			
		\centering
		\includegraphics[width=0.6\textwidth]{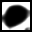}
		\caption{`digit\_9'}
		\label{subfig_miss_classification_3}
	\end{subfigure}
	\begin{subfigure}[t]{0.12\textwidth}			
		\centering
		\includegraphics[width=0.6\textwidth]{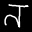}
		\caption{`Ta'}
		\label{subfig_miss_classification_4}
	\end{subfigure}
	~
	\begin{subfigure}[t]{0.12\textwidth}			
		\includegraphics[width=0.6\textwidth]{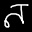}
		\caption{`Na'}
		\label{subfig_miss_classification_5}
	\end{subfigure}
	\caption{Miss-classification analysis: (a) confusion matrix of UCI Devanagari numeral dataset, (b) actual `digit\_0' in UCI Devanagari numeral dataset, (c) and (d) show `digit\_7' and 'digit\_9', respectively, miss-classified as `digit\_0' on UCI Devanagari numeral dataset, (e) and (f) actual `ta' and `na' which is miss-classified as `ta' on IAPR-11 (O) Devanagari dataset.}
	\label{fig_miss-classification}
\end{figure}

 Sub-fig.~\ref{subfig_miss_classification_confusion_mat} presents a confusion matrix which shows only two misclassifications where `digit\_1' and `digit\_7' are classified as `digit\_0'. For `digit\_1', this is due to deviations in the structure by the writer, i.e., due to bad handwriting and for `digit\_7', this appears to be noise in the recording process as some part of the character seems to be cropped, and it is impossible even for humans to find the class of the character.
 
 Sub-figs.~\ref{subfig_miss_classification_4} and \ref{subfig_miss_classification_5} study misclassifications on the IAPR-11 dataset, which is a small dataset. Here, one reason for the misclassifications is due to the similarity in the structure of the characters. As it is clear from the figures, the character `na' is miss-classified as `ta' because they look quite similar, in fact, this is the major cause for misclassifications. Thus, as observed in the literature \cite{Deore2020}, bad handwriting, errors/noises in the recording process and similarity in the structure of characters cause misclassifications.

\section{Conclusion and discussion}
\label{sec_conclusion}
HCR is a widely studied challenging learning problem in pattern recognition, which has a variety of applications, like in the automated processing of documents.
However, there is a lack of research on script independent HCR. This is mainly because of the focus of conventional research on handcrafted feature extraction techniques, the diversity of different scripts and the unavailability of existing datasets and code repositories. Moreover, deep learning, especially CNN, provides a great opportunity to develop script independent models, however, deep learning research in handwriting is still in its infancy and models developed for HCR are focused on specific scripts.

{\color{black}This paper proposed the first script independent deep learning architecture for HCR, called HCR-Net, and started a new research direction for HCR research to develop script independent techniques. HCR-Net uses a novel transfer learning approach for HCR, which partly utilizes a pre-trained VGG16 network to initialize some parts of HCR-Net, unlike the existing techniques which utilize the entire feature extraction layers. The proposed transfer learning technique is based on the hypothesis that HCR is a simpler task as compared to tasks for which pre-trained networks are developed so HCR does not need all the feature extraction layers of the pre-trained networks. Powered by transfer learning and image augmentation, HCR-Net is a computationally efficient technique which can learn faster, and learn on small datasets, unlike standard deep learning techniques which need large amounts of data, and provide better generalizations across several scripts. This work is reproducible, and publicly released at \url{https://github.com/jmdvinodjmd/HCR-Net}.}

The empirical results proved the efficacy of HCR-Net on 40 publicly available datasets of Bangla, Punjabi, Hindi, English, Swedish, Urdu, Farsi, Tibetan, Kannada, Malayalam, Telugu, Marathi, Nepali and Arabic languages. {\color{black}These datasets do not contain any sensitive information about the writers, mitigating privacy concerns.} HCR-Net established 26 new benchmark results while performing close to the best results in the rest cases, and showed performance improvements up to 11\% against the existing results, which presents HCR-Net as a script independent architecture for HCR. HCR-Net also significantly outperformed state-of-the-art transfer learning techniques for HCR and was able to reduce number number of trainable parameters of corresponding VGG16 by 34\%. In addition to that, among the transfer learning techniques, HCR-Net has the fastest convergence rate as it can achieve up to 99\% of final performance in the very first epoch. From miss-classification analysis, it is observed that errors occur mainly due to noisy datasets, bad handwriting and similarity in different characters. {\color{black}We acknowledge that while most of the datasets are recorded in controlled writer conditions, we observed the largest performance improvement with Kannada-mnist, which was collected from practical real-world situations. This indicates that HCR-Net is capable of handling dataset biases and adapting to diverse handwriting styles, even those encountered in real-world scenarios.}

{\color{black}HCR-Net is a promising deep learning technique for HCR, but it has room for improvement, especially in languages with large character sets. In the future, we plan to specialize and extend HCR-Net for these languages, as well as explore hierarchical versions to address the issue of misclassifications due to character similarity. Additionally, as we move towards data-centric AI, we believe there are opportunities to improve HCR by developing specialized pre-processing pipelines and leveraging advanced data-centric methodologies. Aligned with our overarching research thrust towards script independence, we envision the seamless integration of HCR-Net into a comprehensive handwriting recognition system. This integrated system positions handwriting recognition as a pivotal constituent and necessitates nuanced treatment of handwriting independent of scripts and languages. This strategic integration seeks to contribute to the broader landscape of handwriting recognition, emphasizing adaptability across diverse linguistic and script domains.}

\section*{Disclosure statement}
The authors report there are no competing interests to declare.

\section*{Data availability statement}
All the datasets used in the paper are publicly available and the code to reproduce results is released on GitHub at: \url{https://github.com/jmdvinodjmd/HCR-Net}.

\bibliographystyle{plain}
\bibliography{v15}  

\end{document}